\documentclass[10pt,twocolumn,letterpaper]{article}

\usepackage{iccv}
\usepackage{times}
\usepackage{epsfig}
\usepackage{graphicx}
\usepackage{amsmath}
\usepackage{amssymb}
\usepackage{comment}
\usepackage{cuted}
\usepackage[dvipsnames]{xcolor}
\usepackage{booktabs} % for professional tables
\usepackage[ruled, vlined]{algorithm2e}
\SetKwComment{Comment}{$\triangleright$\ }{}
\newlength\savewidth
\usepackage[accsupp]{axessibility}
\usepackage[pagebackref=true,breaklinks=true,letterpaper=true,colorlinks,bookmarks=false]{hyperref}

\iccvfinalcopy % *** Uncomment this line for the final submission

\def\iccvPaperID{11183} % *** Enter the ICCV Paper ID here

\ificcvfinal\fi

\usepackage{amssymb}
\usepackage{amsmath,amsfonts}
\usepackage{amsopn}
\usepackage{bm} % bold symbol
\usepackage{multirow}

\newcommand{\vct}[1]{\boldsymbol{#1}} % vector
\newcommand{\mat}[1]{\boldsymbol{#1}} % matrix
\newcommand{\ProbOpr}[1]{\mathbb{#1}}
\newcommand{\expect}[2]{%
\ifthenelse{\equal{#2}{}}{\ProbOpr{E}_{#1}}
{\ifthenelse{\equal{#1}{}}{\ProbOpr{E}\left[#2\right]}{\ProbOpr{E}_{#1}\left[#2\right]}}} % Expectation: syntax: E{1}{2} = E_1[2], E{}{2}=E[2], E{1}{} = E_1
\newcommand{\var}[2]{%
\ifthenelse{\equal{#2}{}}{\ProbOpr{VAR}_{#1}}
{\ifthenelse{\equal{#1}{}}{\ProbOpr{VAR}\left[#2\right]}{\ProbOpr{VAR}_{#1}\left[#2\right]}}} % Expectation: syntax: V{1}{2} = V_1[2], V{}{2}=V[2], V{1}{} = V_1
\DeclareMathOperator{\argmax}{arg\,max}

\newcommand{\vtheta}{\vct{\theta}}

\newcommand{\vx}{{\vct{x}}}

\newcommand{\vz}{{\vct{z}}}

\newcommand{\vw}{\vct{w}}

\newcommand{\mV}{\mat{V}}

\newcommand{\mJ}{\mat{J}}

\newcommand{\eat}[1]{}
\newcommand{\method}[1]{\textsc{#1}}

\newcommand{\MFW}{\method{MFW}\xspace}

\begin{document}

\title{Procrustean Training for Imbalanced Deep Learning}

\author{Han-Jia Ye \qquad De-Chuan Zhan\\
	\small State Key Laboratory for Novel Software Technology, Nanjing University, China\\
	{\tt\small \{yehj, zhandc\}@lamda.nju.edu.cn}
	\and
	Wei-Lun Chao\\
	\small The Ohio State University, USA\\
	{\tt\small chao.209@osu.edu}
}

\maketitle

\thispagestyle{empty}

%%%%%%%%% ABSTRACT
%!TEX root=main.tex
\begin{abstract}
Neural networks trained with class-imbalanced data are known to perform poorly on minor classes of scarce training data. Several recent works attribute this to over-fitting to minor classes. In this paper, we provide a novel explanation of this issue. We found that a neural network tends to first under-fit the minor classes by classifying most of their data into the major classes in early training epochs. To correct these wrong predictions, the neural network then must focus on pushing features of minor class data across the decision boundaries between major and minor classes, leading to much larger gradients for features of minor classes. We argue that such an under-fitting phase over-emphasizes the competition between major and minor classes, hinders the neural network from learning the discriminative knowledge that can be generalized to test data, and eventually results in over-fitting. To address this issue, we propose a novel learning strategy to equalize the training progress across classes. We mix features of the major class data with those of other data in a mini-batch, intentionally weakening their features to prevent a neural network from fitting them first. We show that this strategy can  largely balance the training accuracy and  feature gradients across classes, effectively mitigating the under-fitting then over-fitting problem for minor class data. On several benchmark datasets, our approach achieves the state-of-the-art accuracy, especially for the challenging step-imbalanced cases.

\end{abstract}

%%%%%%%%% BODY TEXT

%!TEX root=main.tex
\section{Introduction}
\label{s_intro}

\begin{figure}[t!]
	\centering
	\minipage{0.48\linewidth}
	\centering
	\mbox{\small (a) training set accuracy}
	\includegraphics[width=1.\linewidth]{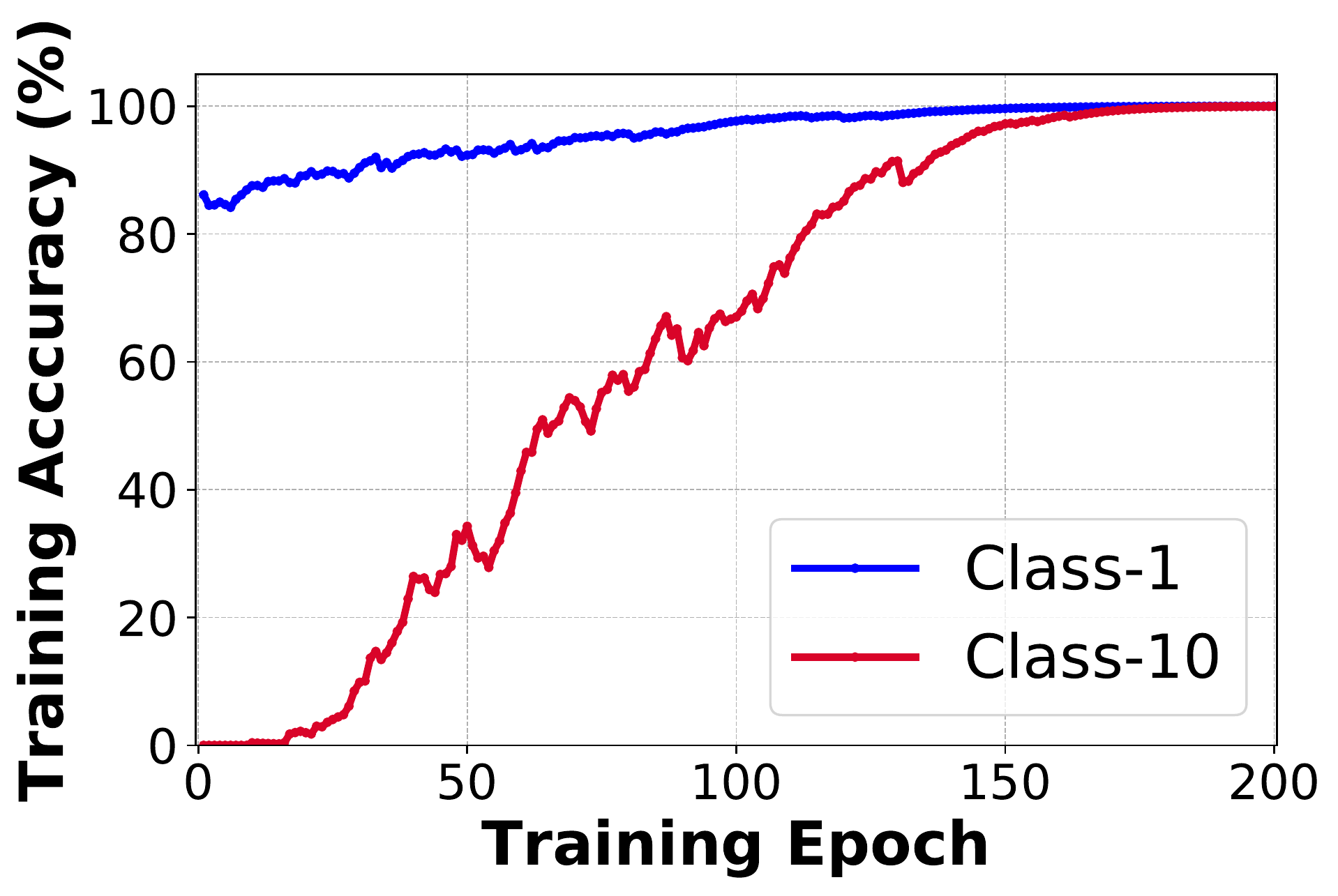}
	\endminipage
	\hfill
	\minipage{0.48\linewidth}
	\centering
	\mbox{\small (b) classification ratio}
	\includegraphics[width=1.\linewidth]{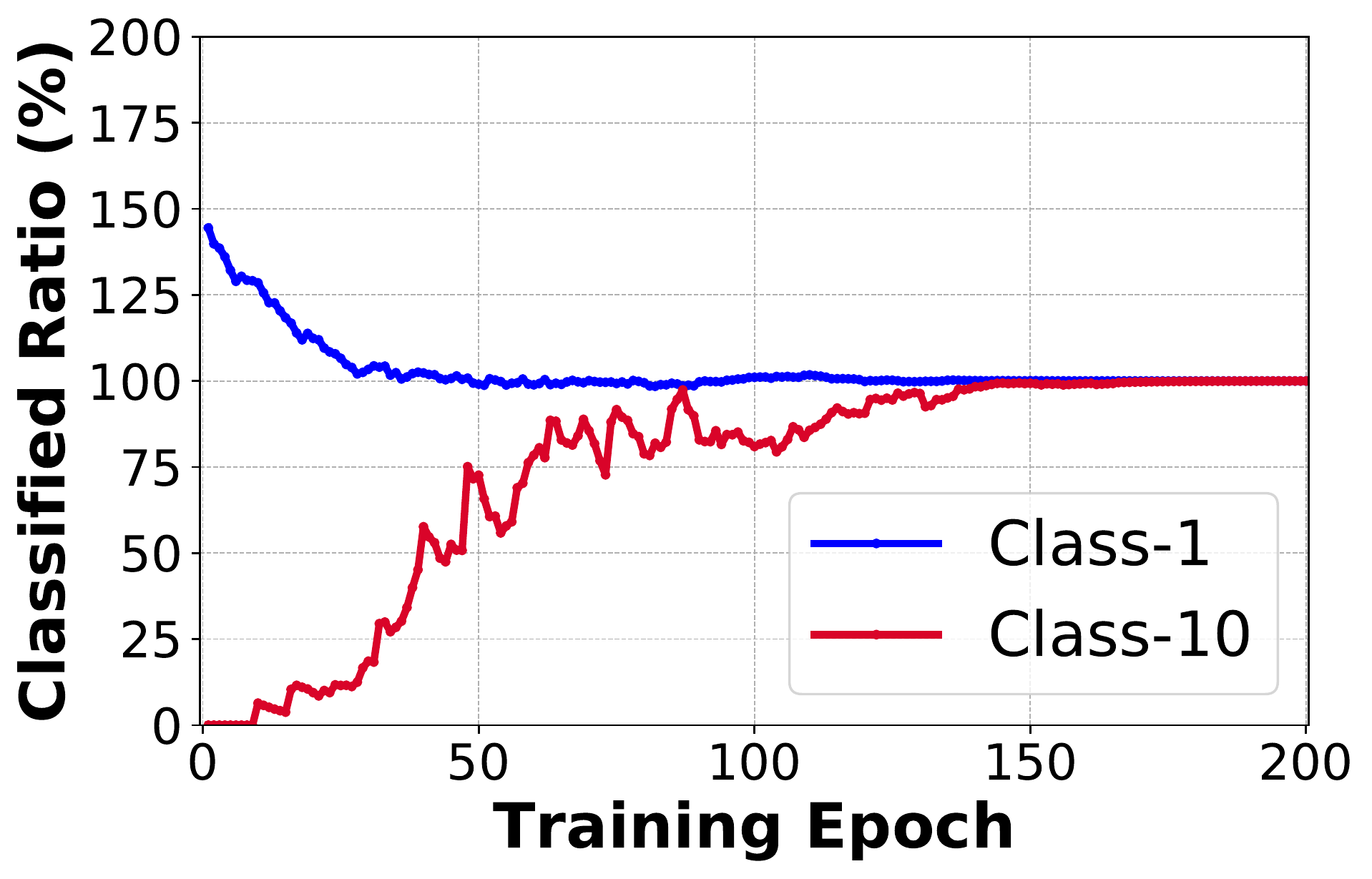}
	\endminipage
	%\vskip 5pt
	\caption{\small \textbf{The \emph{training progress} of a neural network on class-imbalanced data.} We trained a ResNet-32 \cite{he2016deep} on a long-tailed CIFAR-10 data set \cite{krizhevsky2009learning}. We only showed two classes for clarity. Class-1 and class-10 have $5000$ and $50$ training instances, respectively. 
	\textbf{(a) training accuracy per class} along the epochs. \textbf{(b) classification ratio:} the numbers of training instances classified into a class \emph{divided by} the number of training instances which truly belong to that class, along the epochs.
    Minor classes need longer time to achieve almost a $100\%$ training accuracy, and most of their instances are classified into the major classes in early epochs.}
	\label{fig:main-erm_stat}
	\vspace{-10pt}
\end{figure}

Deep neural networks \cite{he2016deep,huang2017densely,krizhevsky2012imagenet,simonyan2015very,szegedy2015going} trained with \emph{class-imbalanced data}, of which we have ample data for some ``major'' classes and limited data for the other ``minor'' classes, are known to perform poorly on minor classes \cite{buda2018systematic,liu2019large,van2017devil}. 
As many real-world data sets are class-imbalanced by nature, especially those for recognizing a large number of objects \cite{guo2016ms,krishna2017visual,lin2014microsoft,thomee2015yfcc100m,van2018inaturalist}, this problem has attracted increasing attention recently in both the machine learning and computer vision communities \cite{cao2019learning,cui2019class,kang2019decoupling,tang2020long,sagawa2019distributionally,wang2021long}.

Many works have attempted to explain this problem and develop corresponding solutions. 
Some attribute this to the mismatch between training and test data distributions \cite{cao2019learning,chawla2002smote,he2009learning,khan2017cost,johnson2019survey,wang2016training,tan2020equalization,Jamal2020Rethinking,ren2020balanced}\footnote{During testing, one usually assumes \emph{class-balanced} test data or computes average per-class accuracy.} or \emph{under-fitting to minor classes} \cite{li2016solving,li2018adaptive}. 
Others attribute this to the poor feature learning \cite{dong2017class,dong2018imbalanced,huang2016learning,huang2019deep,zhang2017range,hayat2019gaussian,wu2017deep,zhong2019unequal} or \emph{over-fitting to minor classes} \cite{al2016algorithms,Ye2020CDT,Kim2020Adjusting}. That is, a neural network can \emph{easily} fit the limited amount of minor class data to obtain $\sim 100\%$ training accuracy, but just cannot generalize to the test data. One particular finding made by~\cite{Ye2020CDT,Kim2020Adjusting} is the \emph{feature deviation} between training and test data: for minor classes, the training and test features generated by the learned neural network\footnote{Here the features mean the neural network's outputs before the last fully-connected layer.} deviate from each other, making the learned decision boundary not applicable to test data.

In this paper, we present a novel finding that links these two lines of explanations.
We analyzed the
\textbf{training progress of a neural network} (\ie, its training accuracy along the epochs; see  \autoref{fig:main-erm_stat}).
We found that, while a neural network eventually fits the minor class data, it only happens when the training is about to converge. 
In fact, in most of the early training process, the neural network classifies most of the minor class training data into major classes, essentially under-fitting the minor class data.  
To correct these wrong predictions, the neural network then must focus on pushing features of the minor class data across the decision boundaries between major and minor classes.

We argue that, this initial \emph{under-fitting phase of minor classes} exaggerates the competition between major and minor classes, forcing the neural network to overly learn the discriminative knowledge that cannot be generalized and ultimately leading to feature deviation and over-fitting.
Concretely, in training the feature extractor $f_{\vtheta} (\vx)$\footnote{$f_{\vtheta}$ is the part of a neural network before the last fully-connected layer.} using a loss function $\ell$, we found that the gradients assigned to features of minor class data (\ie, $\nabla_{f_{\vtheta} (\vx)} \ell$) are much higher than those for major class data (as will be analyzed in \autoref{s_approach}). This finding provides an explanation to feature deviation \cite{Kim2020Adjusting,Ye2020CDT}: the exaggerated gradients push the features of minor class training data further away than where they need to be, thus deviating from the features of the test data.

To address this issue, we propose a novel, simple yet effective learning strategy to alleviate the \emph{under-fitting then over-fitting}
problem for minor class data.
The core idea is to \emph{equalize} the training progress between major and minor classes by suppressing a neural network's tendency to first fit major class data --- hence making it \emph{procrustean across classes}. 
We achieve this by \emph{weakening} the features of major class data at every mini-batch.
Concretely, for a major class datum, we mix its feature with that of another data (\ie, a convex interpolation), such that the resulting features will likely move toward or even across the decision boundary (thus misclassified).
We showed that, this learning strategy can effectively balance not only the training accuracy across classes, but also the gradients assigned to features. The resulting neural network therefore suffers less feature deviation and over-fitting for minor classes.

We validate our approach on several benchmark data sets, including CIFAR-10 \cite{krizhevsky2009learning}, CIFAR-100 \cite{krizhevsky2009learning}, TinyImageNet \cite{le2015tiny}, and iNaturalist \cite{van2018inaturalist}. 
Our approach achieves the state-of-the-art results on many of the experimental settings, \emph{especially for the more  challenging step-imbalanced cases}. By analyzing the learned features, we also observe much smaller feature deviations, essentially resolving one fundamental issue in class-imbalanced deep learning.

%!TEX root=main.tex
\section{Related Work}
\label{s_related}

There are two mainstream methods of class-imbalanced learning: \emph{re-sampling} based and  \emph{cost-sensitive} based. 

\noindent\textbf{Re-sampling based methods} aims to change the training data distribution to match the balanced test data~\cite{drummond2003c4,buda2018systematic,van2017devil}. Exemplar ways are over-sampling the minor-class data~\cite{buda2018systematic,byrd2019what,shen2016relay} or under-sampling the major-class data~\cite{buda2018systematic,he2009learning,japkowicz2000class,ouyang2016factors}, directly from the training data. 
Some methods synthesized additional minor-class data to enlarge the diversity of a class~\cite{AdSampling,chawla2002smote,fernandez2018smote,wang2019wgan}. Some others transferred statistics from the major classes to the minor classes~\cite{hariharan2017low,Kim2020M2M,liu2019large,yin2019feature}. \cite{chou2020remix} proposed Remix to synthesize new data by linearly interpolating two real data. Different from mixup~\cite{zhang2018mixup}, Remix used different mixing coefficients for the input data and labels (larger label coefficients for minor classes), essentially generating more data for minor classes. In contrast, we perform linear interpolation only in the input data. Our goal is to balance the training progress, not to augment more data for minor classes.

\noindent\textbf{Cost-sensitive based methods} adjust the cost of incorrect predictions according to the true class labels.
One popular way is re-weighting, which gives each instance a weight based on its true label when computing the total loss. Setting the weights by (the square roots of) the reciprocal of the number of training instances per class has been widely used~\cite{huang2016learning,huang2017discriminative,mahajan2018exploring,wang2016training,wang2017learning}. \cite{cui2019class} proposed a principled way to set weights by computing the effective numbers of training instances. 
\cite{chang2017active,jiang2018mentornet,ren2018learning,shu2019meta,wang2019dynamic,Jamal2020Rethinking,ren2020balanced} explored dynamically adjusting the weights via meta-learning or curriculum learning.
Instead of adjusting instance weights, \cite{khan2017cost} developed several instance loss functions that reflect class imbalance; \cite{cao2019learning,Ye2020CDT} forced minor-class instances to have large additive or multiplicative margins from the decision boundaries. 
\cite{khan2019striking} proposed to incorporate uncertainty of instances or classes in the loss function. The closest to ours are~\cite{li2020overcoming,tan2020equalization}, which introduced new loss functions to balance the gradients assigned to the last fully-connected layers (\ie, linear classifiers) over classes. In contrast, we aim to balance the gradients assigned to data instances for better feature learning.

\noindent\textbf{Learning feature embeddings} with class-imbalanced data has also been studied, especially for face recognition \cite{dong2017class,dong2018imbalanced,huang2016learning,huang2019deep,zhang2017range}. \cite{hayat2019gaussian,wu2017deep,zhong2019unequal} combined objective functions of classification and embedding learning to better exploit minor class data. \cite{wang2020frustratingly,cui2018large,zhang2019study} proposed two-stage training procedures to pre-train features with imbalanced data and fine-tune the classifier with balanced data; \cite{kang2019decoupling} systematically studied different training strategies for each stage. 
\cite{Zhou2020BBN} introduced bilateral-branch networks to cumulatively make the two-stage transition.
Our work also improves features, by reducing the unfavorable feature deviation~\cite{Ye2020CDT,Kim2020Adjusting}.

\noindent\textbf{Empirical observations.} Similar to ours, several recent works are built upon empirical analysis.
\cite{guo2017one,kang2019decoupling,yin2019feature,Kim2020Adjusting} found that the learned linear classifiers of a ConvNet tend to have larger norms for major classes, and proposed to force similar norms across classes in training or calibrate the norms in testing. \cite{wu2017deep} found that the feature norms of major-class and minor-class instances are different and proposed to regularize it by forcing similar norms. 
\cite{Ye2020CDT,Kim2020Adjusting} both found the phenomenon of feature deviation between the training and test data, especially for minor class data. Our work is different by presenting a novel finding of the network training progress. The observed larger gradients on minor class data offer an explanation for feature deviation.
%!TEX root=main.tex
\section{Approach}
\label{s_approach}

In this section, we introduce our approach, which we called Major Feature Weakening (\MFW). We begin with the basic notation, followed by our algorithm. We then provide analysis of its properties, especially on how it balances feature gradients and the training progress across classes.

\subsection{Background and notation}

We denote a $C$-class neural network classifier by
\begin{align}
\hat{y} = \argmax_{c\in\{1,\cdots,C\}} \vw_c^\top f_{\vtheta} (\vx),
\end{align}
where $\vx$ is the input, $f_{\vtheta}(\cdot)$ is the feature extractor parameterized by $\vtheta$, and $\{\vw_c\}_{c=1}^C$ is the final fully-connected layer for linear classification. The feature extractor $f_{\vtheta}(\cdot)$ can be further decomposed by $h_{\vtheta}\circ g_{\vtheta}(\cdot) = h_{\vtheta}(g_{\vtheta}(\cdot))$, where the output of $g_{\vtheta}$ is the intermediate feature. 

Given a training set $D_\text{tr} = \{(\vx_n, y_n)\}_{n=1}^N$, in which each class $c$ has $N_c$ instances, we normally train the classifier by \textbf{empirical risk minimization (ERM)}, using a loss function $\ell(y, \{\vw_c^\top f_{\vtheta} (\vx) \}_{c=1}^C)$,
\begin{align}
&\min_{\vtheta, \{\vw_c\}_{c=1}^C} \sum_n \ell(y_n, \{\vw_c^\top f_{\vtheta} (\vx_n) \}_{c=1}^C) \nonumber\\
=&\min_{\vtheta, \{\vw_c\}_{c=1}^C} \sum_n \ell(y_n, \{\vw_c^\top h_{\vtheta}(g_{\vtheta}(\vx_n)) \}_{c=1}^C).
\label{eq:loss}
\end{align}
One popular loss function is the cross-entropy loss,
\begin{align}
\ell(y, \{\vw_c^\top f_{\vtheta} (\vx) \}_{c=1}^C) = & -\log p(y|\vx; \vtheta, \{\vw_c\}_{c=1}^C) \nonumber \\ = &  -\log\frac{\exp(\vw_{y}^\top f_{\vtheta}(\vx))}{\sum_c \exp(\vw_c^\top f_{\vtheta}(\vx))}. 
\label{eq:CE}
\end{align}
We apply stochastic gradient descent (SGD) for optimization, with uniformly sampled instances from $D_\text{tr}$. 

For class-imbalanced learning, each class $c$ will have a different number of training instances $N_c$.

%%%%
\subsection{Major feature weakening (\MFW)}
\label{ss_MFW}
As mentioned in~\autoref{s_intro} and \autoref{fig:main-erm_stat}, a neural network trained with class-imbalanced data tends to fit major classes first, resulting in an inconsistent training progress across classes. To resolve this problem, we propose to weaken the features of major classes within each mini-batch. 

Let $(\vx_1, y_1)$ and $(\vx_2, y_2)$ be two training data instances in a mini-batch, \MFW performs the following operation to the \underline{intermediate feature $g_{\vtheta}(\vx_1)$ of $\vx_1$}
\begin{align}
   & \tilde{\vz}_1 = (1-\lambda_1)\times g_{\vtheta}(\vx_1) + \lambda_1 \times g_{\vtheta}(\vx_2), \nonumber \\
   & \tilde{y}_1 = y_1, \label{e_MFW}
\end{align}
which mixes (\ie, convexly interpolates) the intermediate feature $g_{\vtheta}(\vx_1)$ with $g_{\vtheta}(\vx_2)$ to become the new intermediate feature $\tilde{\vz}_1$ of $\vx_1$. The label of $\vx_1$ is kept intact. Thus, when $y_2\neq y_1$, \autoref{e_MFW} essentially moves $g_{\vtheta}(\vx_1)$ toward features of other classes, hence \emph{weakening} its feature.
The resulting $(\tilde{\vz}_1, \tilde{y}_1)$ is then fed into $h_{\vtheta}$ to obtain the feature of $\vx_1$ and calculate the loss.

Here, $\lambda_1\in[0, 1]$ is sampled from a beta distribution $\text{Beta}(\alpha, \alpha)$ following \cite{zhang2018mixup}, which is then multiplied with a class-dependent weight $s(N_{y_1})$. The weight function $s(\cdot)$ is monotonically increasing with the class size $N_{y_1}$ and has a range $[0, 0.5]$, which gives major classes a larger weight to weaken their features. 
That is, $g_{\vtheta}(\vx_1)$ will be weakened more with a larger $\lambda_1$ if $y_1$ is a major class.
Nevertheless, the range of $s$ and the support of the beta distribution ensure that $g_{\vtheta}(\vx_1)$ is still the main ingredient of the new intermediate feature $\tilde{\vz}_1$. 
Algorithm \ref{a_MFW} summarizes the  training procedure of \MFW. We discuss how to set $s(\cdot)$ in~\autoref{eq:lambda_n}.

During evaluation, given a training or test example $\vx$, we do not perform \MFW but extract its feature by $h_{\vtheta}(g_{\vtheta}(\vx))$.

%%%%%%%%%%%%%%%%%
\setlength{\textfloatsep}{8pt}
\begin{algorithm}[t]
\SetAlgoLined
\label{a_MFW}
\caption{Major Feature Weakening (\MFW): see~\autoref{ss_MFW} for details.}
\SetKwInOut{SInput}{Input}
\SetKwInOut{SModel}{Model}
\SInput{training data $D_\text{tr} = \{(\vx_n, y_n)\}_{n=1}^N$; initial parameters $\vtheta$, $\{\vw_c\}_{c=1}^C$; weight function $s$; beta distribution coefficient $\alpha$; batch size $B$}
\SModel{$f_{\vtheta} = h_{\vtheta}\circ g_{\vtheta}$}
\While{not converge}{
\textbf{Sample} $D_1 = \{(\vx^{(1)}_n, y^{(1)}_n)\}_{n=1}^B$ from $D_\text{tr}$ \\
\textbf{Permute} $D_1$ to get $D_2 = \{(\vx^{(2)}_n, y^{(2)}_n)\}_{n=1}^B$ \\
\For{$n\in\{1,\cdots,B\}$}{
$\lambda_n \sim \text{Beta}(\alpha, \alpha)$\\
$\lambda_n \leftarrow {\color{blue}s(N_{y^{(1)}_n})} \times \lambda_n$\\
${\color{blue}\tilde{\vz}_n} = (1-\lambda_n)\times g_{\vtheta}(\vx^{(1)}_n) + \lambda_n \times g_{\vtheta}(\vx^{(2)}_n) $\\
$\tilde{y}_n = y_n^{(1)}$\\
}
\textbf{Optimize} \autoref{eq:loss} w.r.t. $\vtheta$ and $\{\vw_c\}_{c=1}^C$ using $\tilde{D} = \{({\color{blue}\tilde{\vz}_n}, \tilde{y}_n)\}_{n=1}^B$,where $g_{\vtheta}(\vx_n)$ in \autoref{eq:loss} is replaced by ${\color{blue}\tilde{\vz}_n}$.
}
\end{algorithm}

%%%%%%%%%%%%%%%%%

\begin{figure*}[t!]
		\centering
		\includegraphics[width=0.95\linewidth]{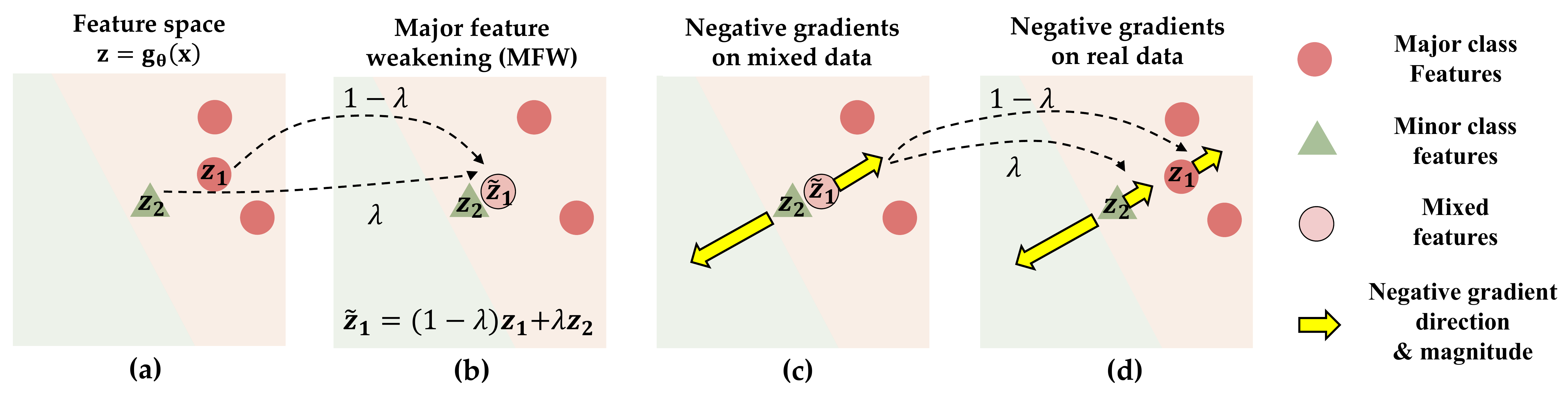}
	\vskip -5pt
	\caption{\small \textbf{Illustration of \MFW.} (a) intermediate features $\vz=g_{\vtheta}(\vx)$; (b) mixed, weakened features $\tilde{\vz}_1$ for a major class; (c) the gradients $\nabla_{\vz_2}\ell$ and $\nabla_{\tilde{\vz}_1}\ell$; (d) the gradients $\nabla_{g_{\vtheta}(\vx_2)}\ell$ and $\nabla_{g_{\vtheta}(\vx_1)}\ell$, where $\vx_2$ and $\vx_1$ are the pre-images of $\vz_2$ and $\vz_1$. 
	From (c) to (d), $\nabla_{\tilde{\vz}_1}\ell$ is separated into two parts, one of them will affect $\nabla_{g_{\vtheta}(\vx_2)}\ell$.
	With \MFW, $\|\nabla_{g_{\vtheta}(\vx_2)}\ell\|_2$ (\ie, to minor class data) can be reduced.}
	\label{supp-fig:illustration}
	\vspace{-10pt}
\end{figure*}

\subsection{Why does \MFW help imbalanced learning?}
\label{ss_MFW_explain}
We now analyze why \MFW can reduce the gradients assigned to features of minor classes and balance the training progress across classes. Without loss of generality, let us consider a binary classification problem, with $c=1$ as the major class and $c=0$ as the minor class. The classifier can be simplified as $\hat{y} = \frac{1}{2}[\text{sign}(\vw^\top f_{\vtheta} (\vx)) + 1]$. The cross-entropy loss for a data instance $(\vx,y)$ thus becomes
\begin{align}
\ell = & -y \times \log \sigma(\vw^\top f_{\vtheta}(\vx)) \nonumber \\ 
& - (1-y)\times \log(1-\sigma(\vw^\top f_{\vtheta}(\vx))), 
\end{align}
where $\sigma(\vw^\top f_{\vtheta}(\vx)) = \cfrac{1}{1 + \exp(-\vw^\top f_{\vtheta}(\vx))}$.

\noindent\textbf{Reducing the gradients.}
Let $(\vx_1, y_1 = 1)$ and $(\vx_2, y_2 = 0)$ be the two data instances in a mini-batch of size two, where $\vx_1$ is from the major class and $\vx_2$ is from the minor class.
Following Algorithm~\ref{a_MFW}, we construct $(\tilde{\vz}_1, \tilde{y}_1 = y_1 = 1)$ and $(\tilde{\vz}_2, \tilde{y}_2 = y_2 = 0)$,
\begin{align}
   & \tilde{\vz}_1 = (1-\lambda_1)\times g_{\vtheta}(\vx_1) + \lambda_1 \times g_{\vtheta}(\vx_2), \nonumber \\
   & \tilde{\vz}_2 = (1-\lambda_2)\times g_{\vtheta}(\vx_2) + \lambda_2 \times g_{\vtheta}(\vx_1). \label{e_back_pp}
\end{align}
\underline{Let us first consider $h_{\vtheta}$ as an identify function: \ie, $f_{\vtheta} = g_{\vtheta}$.}  When \MFW is not applied (\ie, $\lambda_1=\lambda_2=0$), we have
\begin{align}
\nabla_{g_{\vtheta}(\vx_1)} \ell = (\sigma(\vw^\top g_{\vtheta}(\vx_1)) - y_1 )\times\vw, \nonumber\\
\nabla_{g_{\vtheta}(\vx_2)} \ell = (\sigma(\vw^\top g_{\vtheta}(\vx_2)) - y_2)\times\vw, \label{e_noMFW}
\end{align}
where misclassified data have larger gradients.

When \MFW is applied and we have a weight function that gives $c=1$ a weight $0.5$ and $c=0$ a weight $0$ (so major classes have larger weights), we have $\lambda_1\in[0, 0.5]$ while $\lambda_2=0$.
 This leads to
\begin{align}
\nabla_{\tilde{\vz}_1} \ell = (\sigma(\vw^\top \tilde{\vz}_1) - y_1)\times\vw, \nonumber\\
\nabla_{\tilde{\vz}_2} \ell = (\sigma(\vw^\top \tilde{\vz}_2) - y_2)\times\vw, 
\end{align} 
which, by passing the gradients back to $g_{\vtheta}(\vx_1)$ and $g_{\vtheta}(\vx_2)$ according to \autoref{e_back_pp} (note that, we set $\lambda_2=0$ already), then gives us
\begin{align}
\nabla_{g_{\vtheta}(\vx_1)} \ell = & {\color{blue}(1-\lambda_1)} \times (\sigma(\vw^\top {\color{blue}\tilde{\vz}_1}) - y_1)\times\vw, \nonumber\\
\nabla_{g_{\vtheta}(\vx_2)} \ell = & (\sigma(\vw^\top g_{\vtheta}(\vx_2)) - y_2)\times\vw \nonumber \\
 & +{\color{blue}\lambda_1} \times (\sigma(\vw^\top {\color{blue}\tilde{\vz}_1}) - y_1)\times\vw. 
\end{align} 
The second part of  $\nabla_{g_{\vtheta}(\vx_2)} \ell$ comes from $g_{\vtheta}(\vx_2)$ being used to weaken $g_{\vtheta}(\vx_1)$. \autoref{supp-fig:illustration} gives an illustration.

Now suppose $\vx_2$ is not classified correctly by the current model, \ie $\sigma(\vw^\top g_{\vtheta}(\vx_2))>0.5$, we have
\begin{align}
& |(\sigma(\vw^\top g_{\vtheta}(\vx_2)) - y_2)| \geq \nonumber \\
& |(\sigma(\vw^\top g_{\vtheta}(\vx_2)) - y_2) + {\color{blue}\lambda_1} \times (\sigma(\vw^\top {\color{blue}\tilde{\vz}_1}) - y_1)| \geq 0, \nonumber
\end{align}
which means the norm of $\nabla_{g_{\vtheta}(\vx_2)} \ell$ will be reduced by \MFW\footnote{One can show this by plugging in $y_1 = 1$ and $y_2=0$, and consider $(\sigma(\vw^\top g_{\vtheta}(\vx_2))-0)>0.5$ and $\lambda_1 \times (\sigma(\vw^\top {\tilde{\vz}_1})-1)\in[-0.5, 0.0]$.} in comparison to \autoref{e_noMFW}.

\noindent\textbf{Balancing the training progress.} We now analyze the gradient w.r.t. the linear classifier $\vw$.
Without \MFW, it is
\begin{align}
\nabla_{\vw} \ell = & ( \sigma(\vw^\top g_{\vtheta}(\vx_1))-y_1)\times g_{\vtheta}(\vx_1) \nonumber\\
& + (\sigma(\vw^\top g_{\vtheta}(\vx_2)) - y_2)\times g_{\vtheta}(\vx_2). \label{eq_w_ERM}
\end{align}
With \MFW (but $\lambda_2=0$), the gradient w.r.t. $\vw$ becomes
\begin{align}
\nabla_{\vw} \ell = & ( \sigma(\vw^\top \tilde{\vz}_1)-y_1)\times {\color{blue}\tilde{\vz}_1} \nonumber\\
& + (\sigma(\vw^\top g_{\vtheta}(\vx_2))-y_2)\times g_{\vtheta}(\vx_2) \label{eq_w_MFW}\\
= & (\sigma(\vw^\top \tilde{\vz}_1)-y_1)\times \left({\color{blue}(1-\lambda)}g_{\vtheta}(\vx_1) + {\color{blue}\lambda} g_{\vtheta}(\vx_2)\right) \nonumber\\
& + (\sigma(\vw^\top g_{\vtheta}(\vx_2))-y_2)\times g_{\vtheta}(\vx_2). \nonumber
\end{align}
By comparing the first term in \autoref{eq_w_ERM} and \autoref{eq_w_MFW}, \MFW reduces the tendency of $\vw$ to fit major class data\footnote{The 1st term moves $\vw$ toward $(1-\lambda)g_{\vtheta}(\vx_1) + \lambda g_{\vtheta}(\vx_2)$, not $g_{\vtheta}(\vx_1)$.}. 
In other words, besides weakening the features of major class training data,
\MFW also weakens their classifiers. Both can essentially balance the training progress across classes.

\noindent\textbf{Further discussions.} The gradient reduction for $g_{\vtheta}(\vx_2)$ is governed by $\lambda_1$ that is affected by the class size of $y_1$. In theory, most of the data in a mini-batch come from major classes. Thus, very likely a minor class datum will be paired with a major class datum to get its gradient reduced. 

\noindent\textbf{For other $h_{\vtheta}$, and for $\lambda_2\neq 0$.} When $h_{\vtheta}$ is a linear mapping $\mV$, the above conclusions still hold: the only difference is that $\mV$ will be multiplied into the gradients.  
When $h_{\vtheta}$ is a non-linear function, Jacobian matrices are needed for analysis. See the supplementary material for details.
In practice, we found that even with a complex $h_{\vtheta}$ (\eg, several residual network blocks~\cite{he2016deep}), \MFW still effectively improves class-imbalanced learning. 
We note that when $\lambda_2\neq 0$, $\nabla_{g_{\vtheta}(\vx_1)} \ell$ will be affected by $\lambda_2\times(- \sigma(\vw^\top g_{\vtheta}(\vx_2)) - y_2)\times\vw$.

\begin{figure}[t!]
\centering
	\begin{minipage}{0.49\linewidth}
		\centering
		\mbox{\small (a) ERM}
		\includegraphics[width=0.935\linewidth]{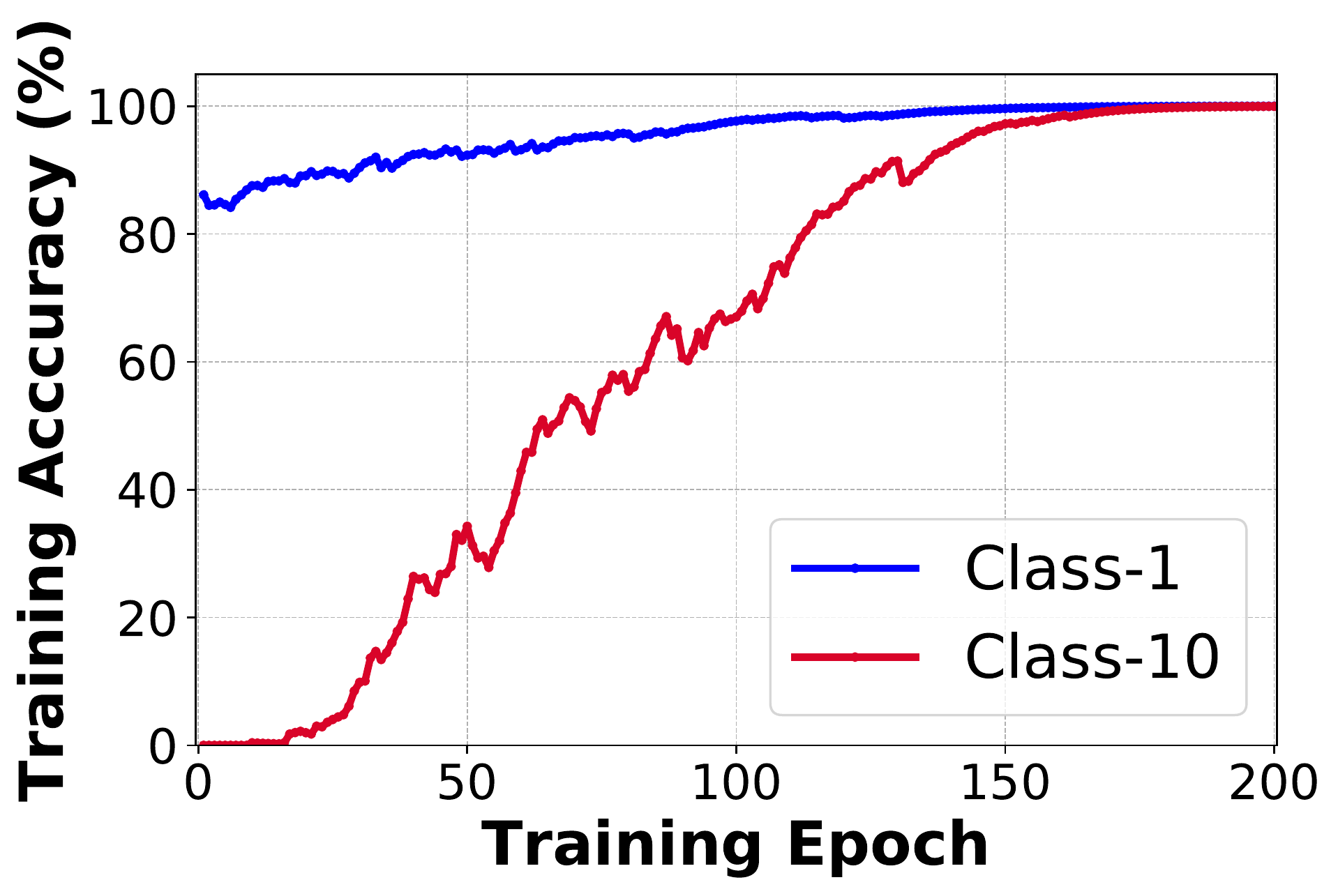}
	\end{minipage}
	\begin{minipage}{0.49\linewidth}
		\centering
		\mbox{\small (b) \MFW}
		\includegraphics[width=0.935\linewidth]{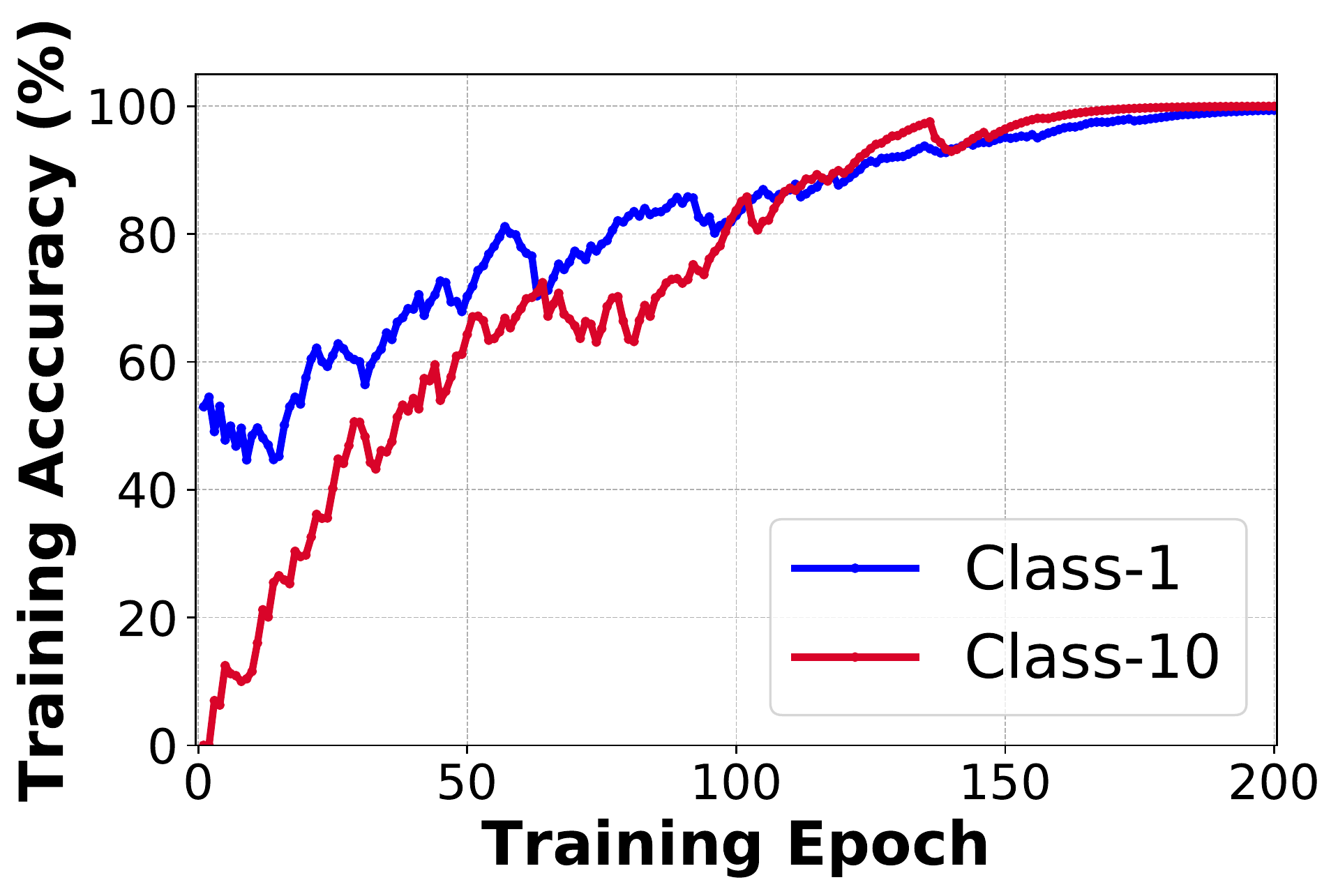}
	\end{minipage}
	\caption{\small \textbf{Training set accuracy by learning with ERM (a) and \MFW (b).} \MFW makes the training progress of the major (\ie, Class-1) and minor (\ie, Class-10) classes more procrustean.}
	\label{fig:train_acc}
	%\vspace{-5pt}
\end{figure}

\begin{figure}[t!]
\centering
	\begin{minipage}{0.49\linewidth}
		\centering
	\mbox{\small (a) $\|\nabla_ {g_{\vtheta}(\vx)}\ell\|_2$ by ERM}
	\includegraphics[width=\linewidth]{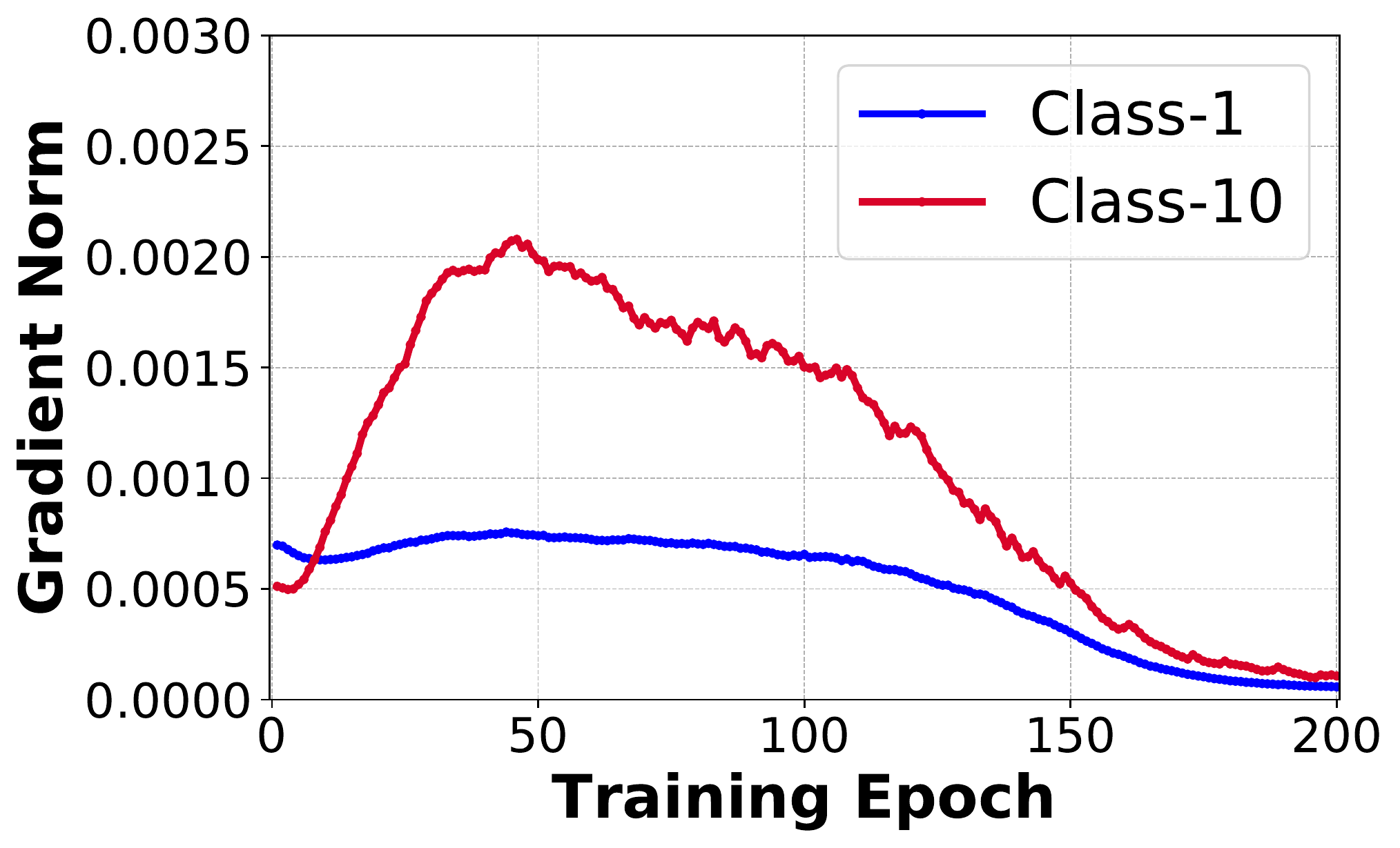}
	\end{minipage}
	\begin{minipage}{0.49\linewidth}
		\centering
	\mbox{\small (b) $\|\nabla_ {g_{\vtheta}(\vx)}\ell\|_2$ by \MFW}
	\includegraphics[width=\linewidth]{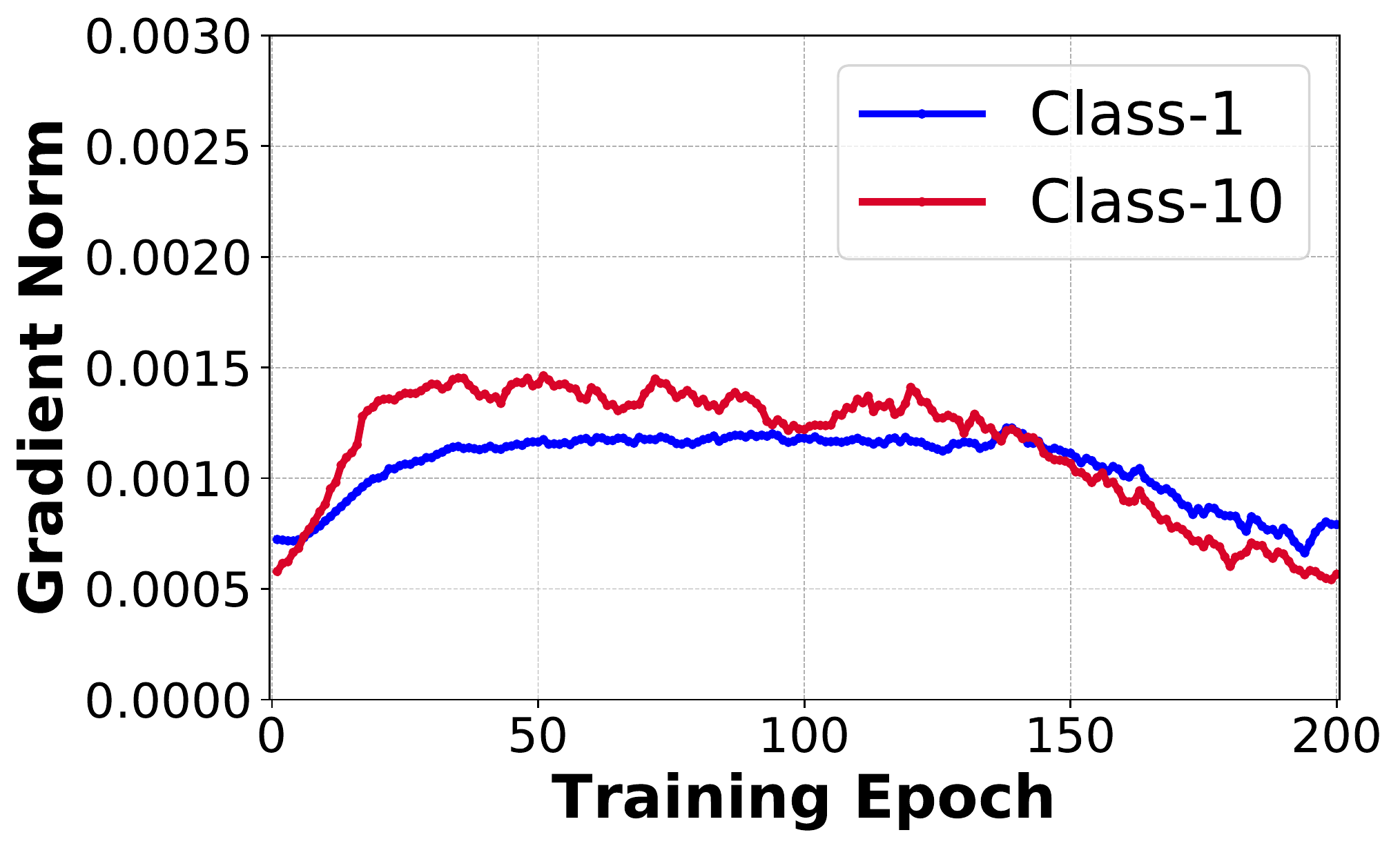}
	\end{minipage}
	\caption{\small \textbf{Feature gradient norm $\|\nabla_ {g_{\vtheta}(\vx)}\ell\|$ by learning with ERM (a) and \MFW (b).} We show $\|\nabla_ {g_{\vtheta}(\vx)}\ell\|$ after every training epoch, averaged over samples of each class. Through \MFW, the gradient norm of the minor class (\ie, Class-10) features decreases; gradient norms across classes are more balanced.}
	\label{fig:gradient}
	%\vspace{-5pt}
\end{figure}

\subsection{Balanced training progress and gradients}
\label{ss_balanced_progress}
We apply \MFW to the same problem as in~\autoref{fig:main-erm_stat}. We train a classifier using ResNet-32~\cite{he2016deep} on a long-tailed CIFAR-10 data set~\cite{krizhevsky2009learning}. The most major class $c=1$ has $5,000$ training examples while the most minor class $c=10$ has $50$ examples. We apply \MFW after the second group of convolutional layers of the ResNet, using a sigmoid-shaped weight function $s(\cdot)$ such that $s(N_1)\approx 0.5$ and $s(N_{10})\approx 0$ (see~\autoref{eq:lambda_n}). Please see \autoref{s_exp} for more details.

\autoref{fig:train_acc} shows the training set accuracy: we evaluate this after each epoch, without altering the features. We include only the two extreme classes for clarity. With \MFW, the accuracy is more balanced across classes. By comparing the gradient norm $\|\nabla_ {g_{\vtheta}(\vx)}\ell\|$
averaged over samples per class in \autoref{fig:gradient}, we see that {learning without \MFW (a)}
has a larger gap of gradient norms between classes, whereas
{learning with \MFW (b)} reduces the gap notably.

We note that, \MFW reduces the gradients given the same network parameters.
This does not imply that \MFW has a smaller gradient norm of minor classes than ERM throughout the entire training process. Indeed, as will be seen in \autoref{s_toy}, \MFW has the effect of keeping samples not too far away from the decision boundary to prevent over-fitting. This means that the training loss at the final training epochs will be larger than ERM, leading to slightly larger gradients.

\subsection{Comparison to mixup and Remix}
\label{ssec:cmp_mixup}
Our \autoref{e_MFW} is reminiscent of mixup~\cite{zhang2018mixup} and~\cite{verma2019manifold}, but with a notable difference: we do not mix the label. Thus, our work does not intend to regularize the neural network to favor a simple linear behavior between training examples.

A recent work Remix \cite{chou2020remix} proposed to use mixup for imbalanced learning by allowing the mixing coefficients of data and labels to be disentangled. Specifically, a higher label mixing coefficient is assigned to minor classes. Their method thus can be seen as re-sampling or data augmentation for minor classes: increasing the minor class examples with linearly interpolated data. In contrast, \MFW does not change the class distribution and hence is not a re-sampling method. Besides, for minor class data, \MFW tends to perform no mixing (no feature weakening). Therefore, \MFW is hardly a data augmentation method for minor classes, but an effective and mathematically sound way to balance the training progress and gradient norms across classes.

  %!TEX root=main.tex
\section{Illustrative experiments}
\label{s_toy}

\begin{figure}[t!]
\centering
    %\text{Training data}
	\begin{minipage}{0.24\linewidth}
		\centering
		\mbox{\small Epoch 20}
		\includegraphics[width=\linewidth]{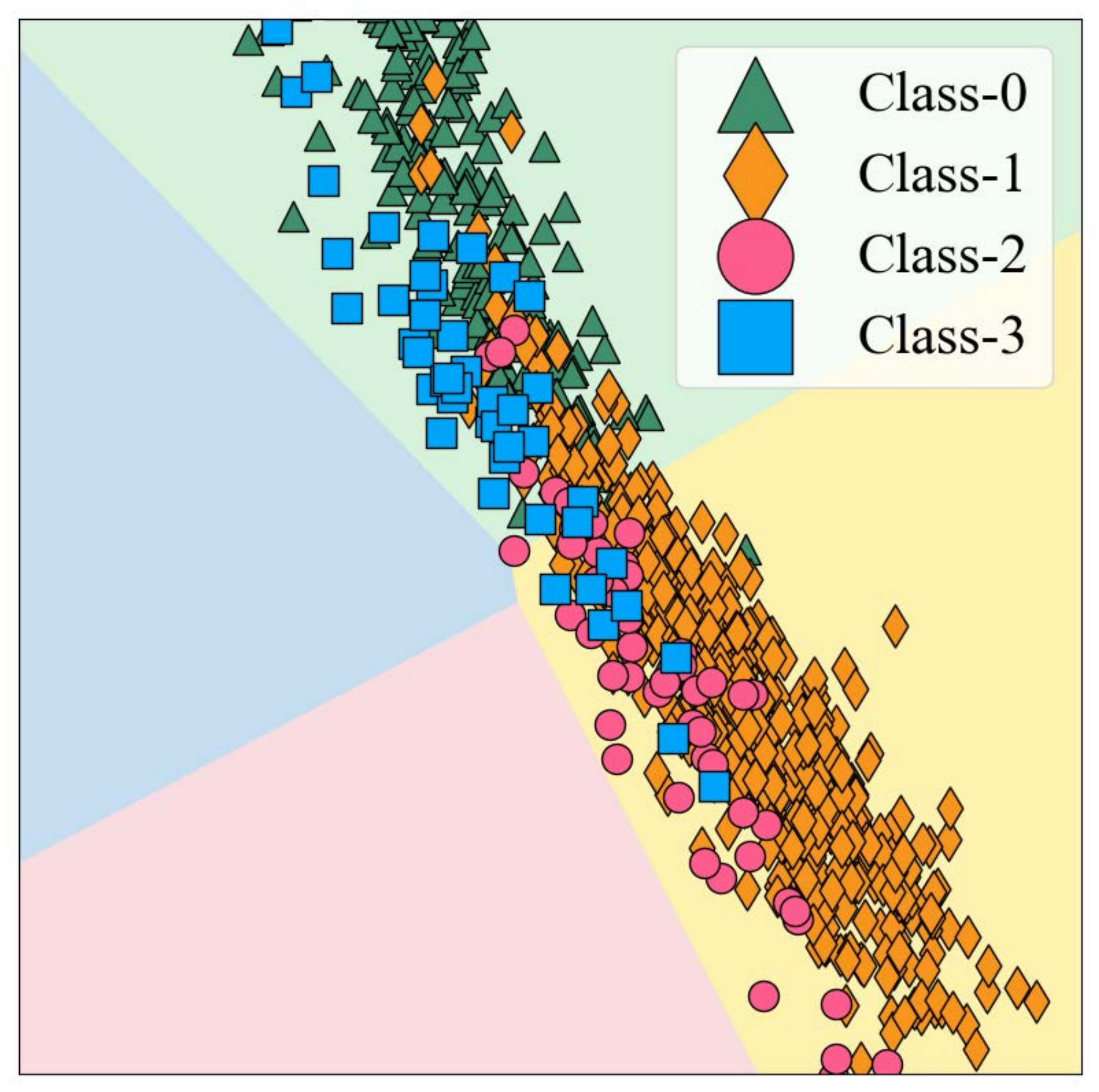}
	\end{minipage}
	\begin{minipage}{0.24\linewidth}
		\centering
		\mbox{\small Epoch 60}
		\includegraphics[width=\linewidth]{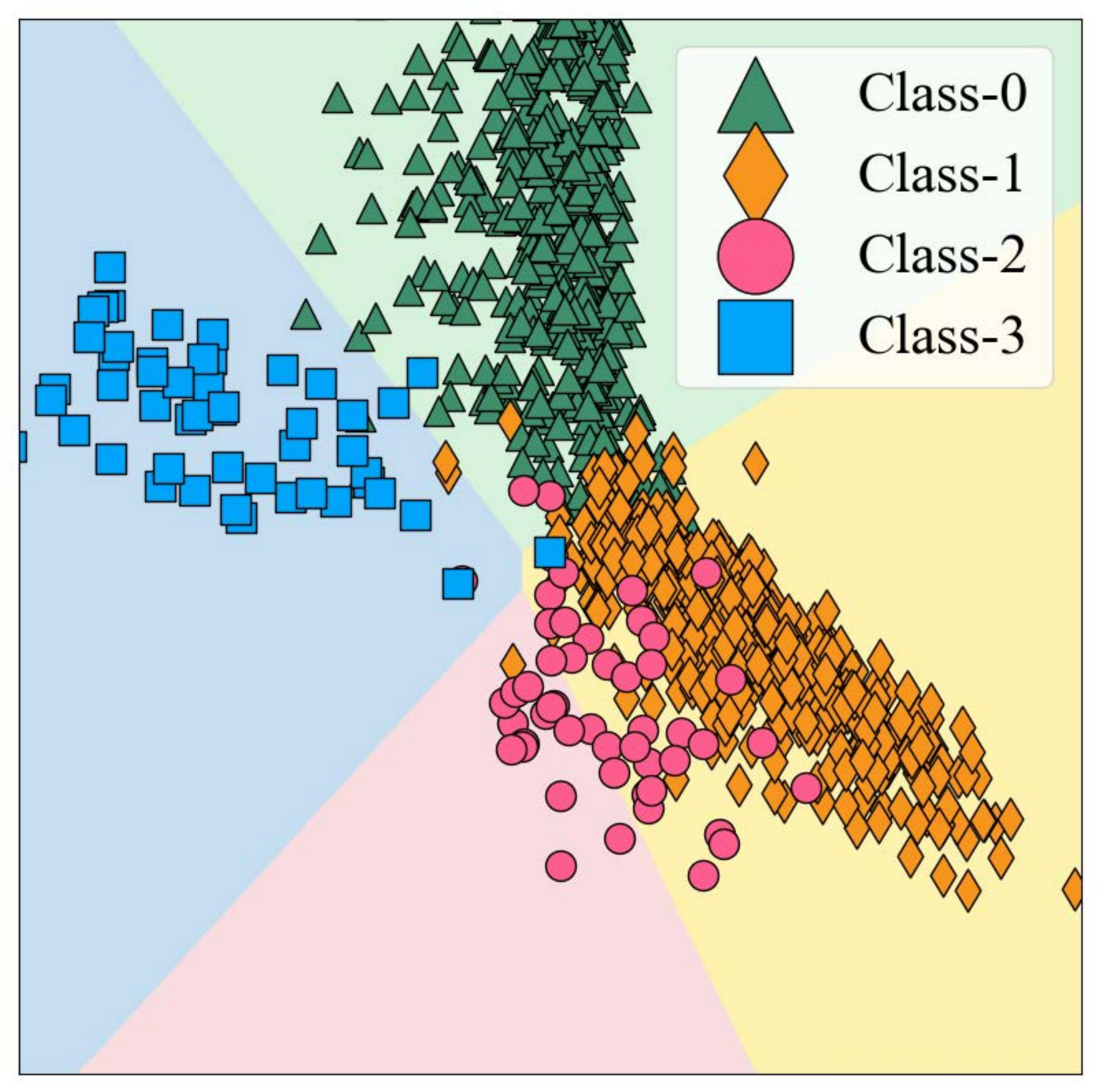}
	\end{minipage}
	\begin{minipage}{0.24\linewidth}
		\centering
		\mbox{\small Epoch 120}
		\includegraphics[width=\linewidth]{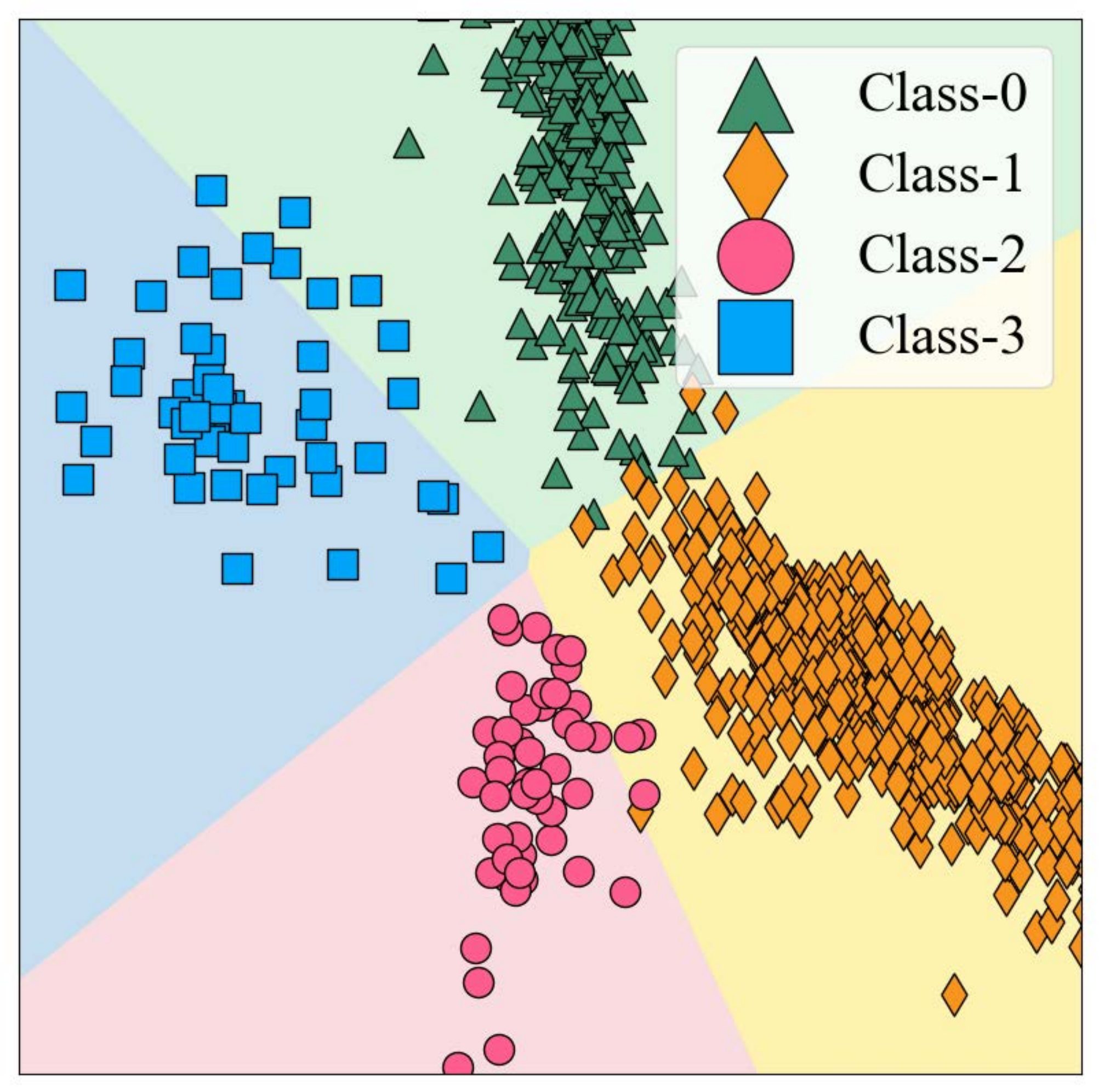}
	\end{minipage}
	\begin{minipage}{0.24\linewidth}
		\centering
		\mbox{\small Epoch 160}
		\includegraphics[width=\linewidth]{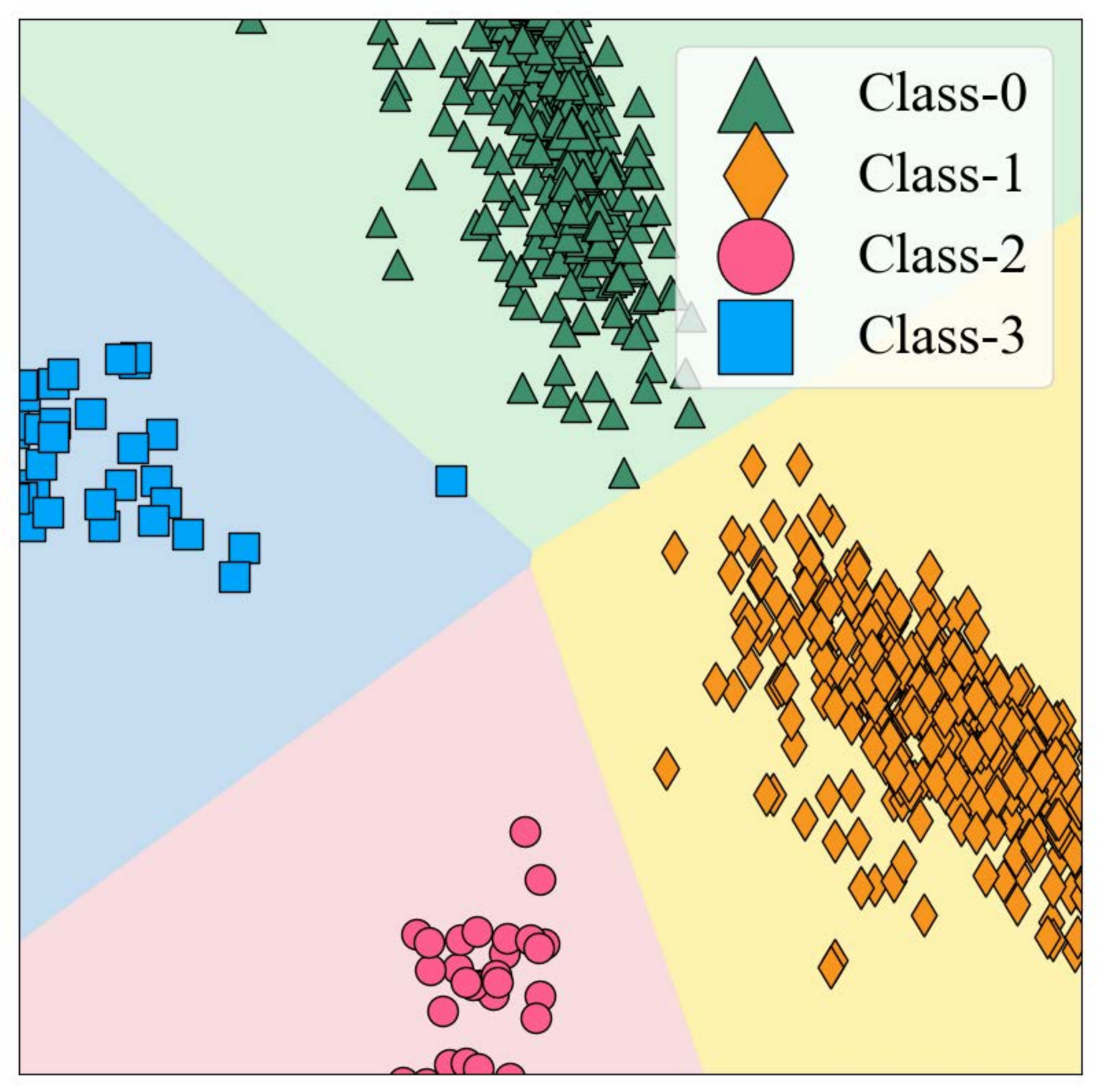}
	\end{minipage}\\
	%\text{Test data} \hspace{15pt}
	\begin{minipage}{0.24\linewidth}
		\centering
		%\mbox{\small Epoch 20}
		\includegraphics[width=\linewidth]{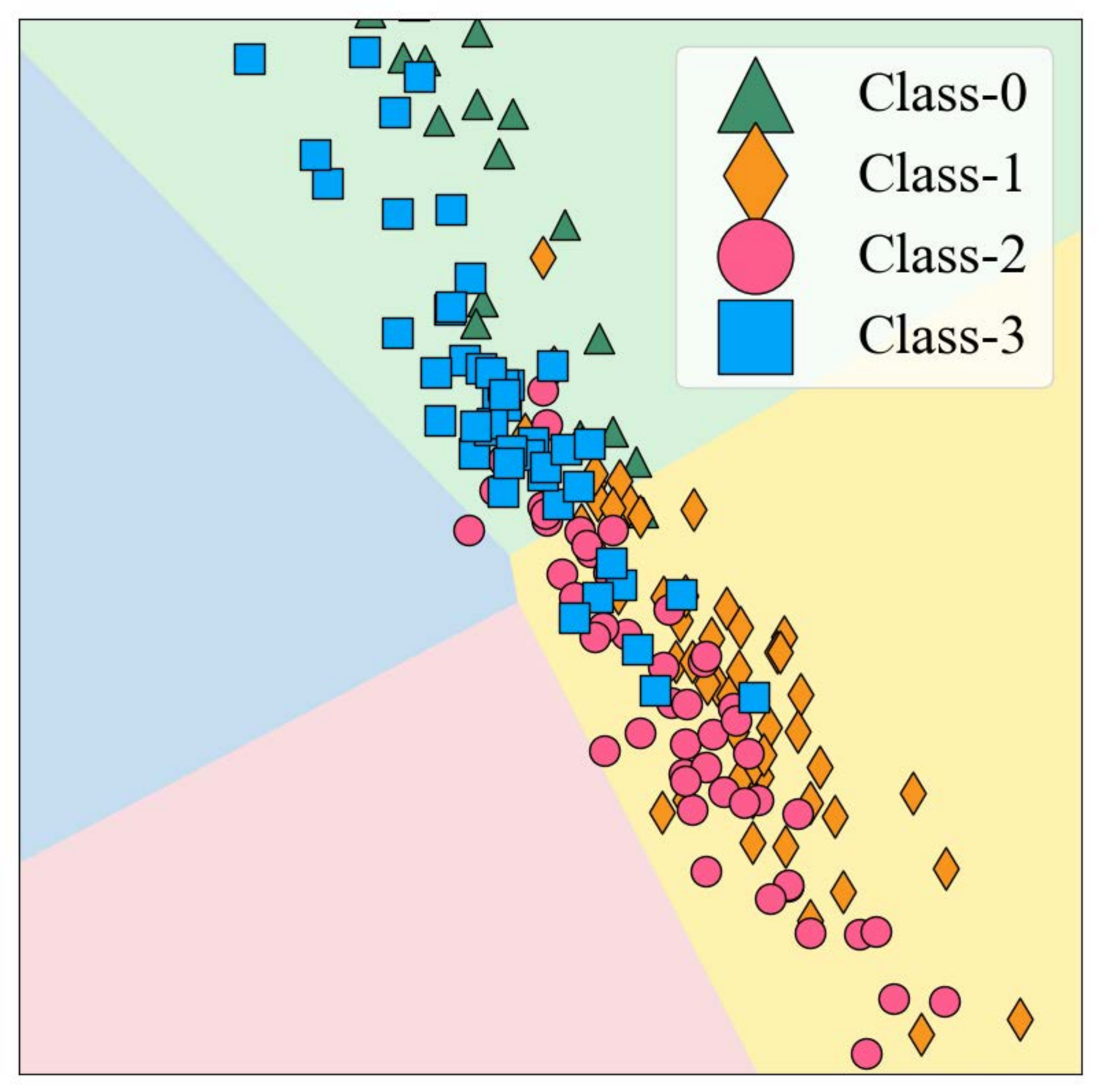}
	\end{minipage}
	\begin{minipage}{0.24\linewidth}
		\centering
		%\mbox{\small Epoch 60}
		\includegraphics[width=\linewidth]{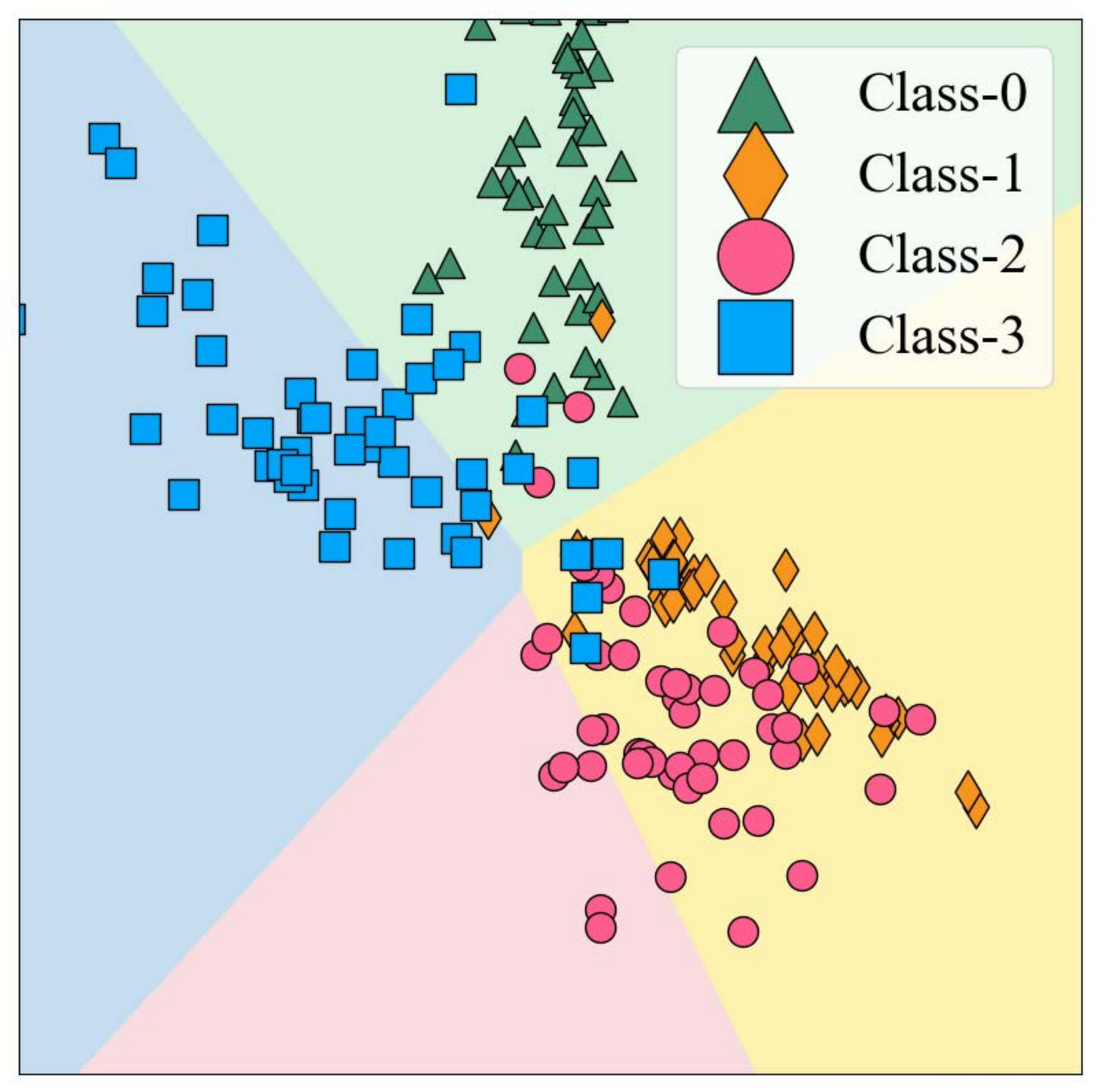}
	\end{minipage}
	\begin{minipage}{0.24\linewidth}
		\centering
		%\mbox{\small Epoch 120}
		\includegraphics[width=\linewidth]{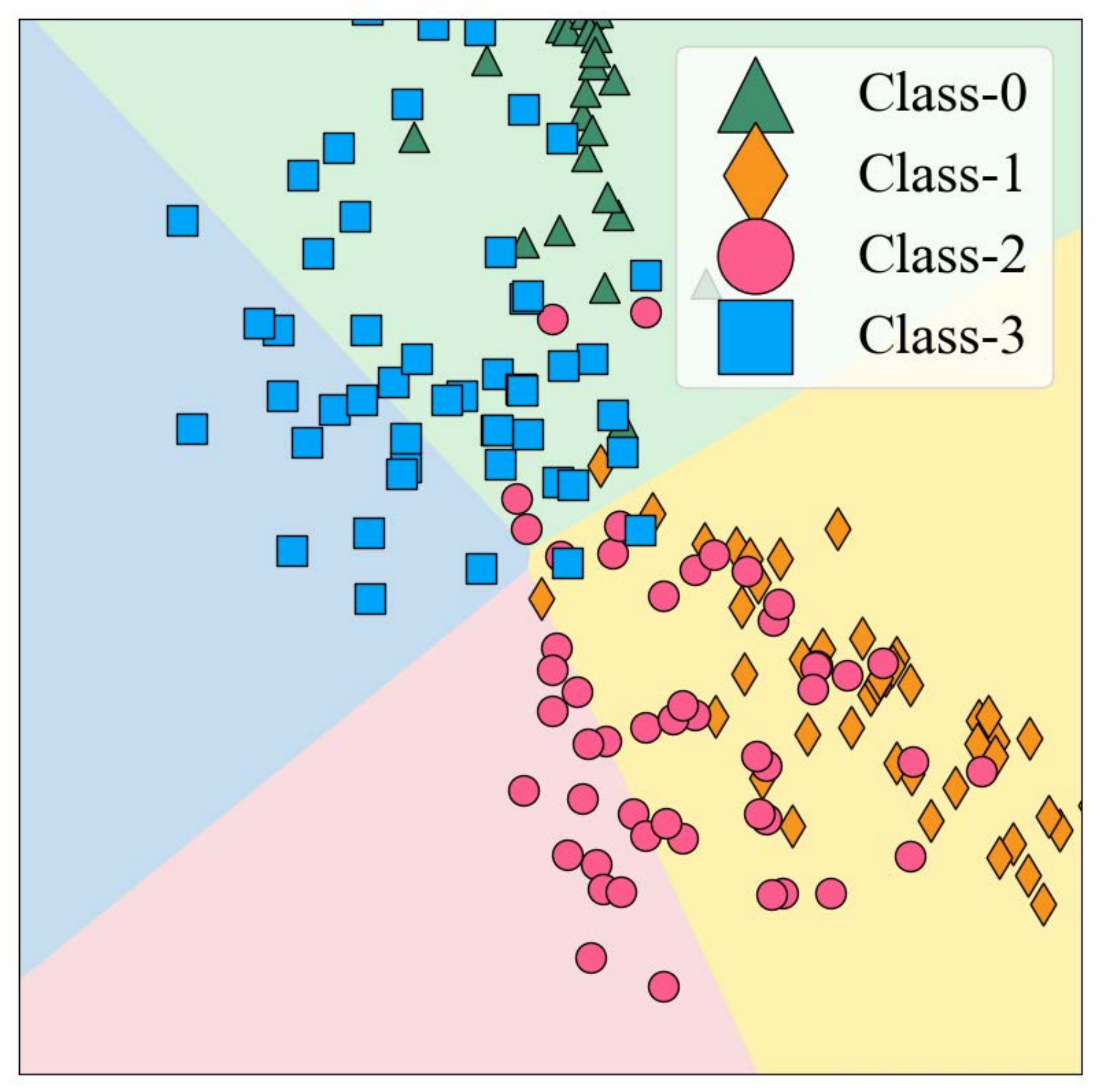}
	\end{minipage}
	\begin{minipage}{0.24\linewidth}
		\centering
		%\mbox{\small Epoch 160}
		\includegraphics[width=\linewidth]{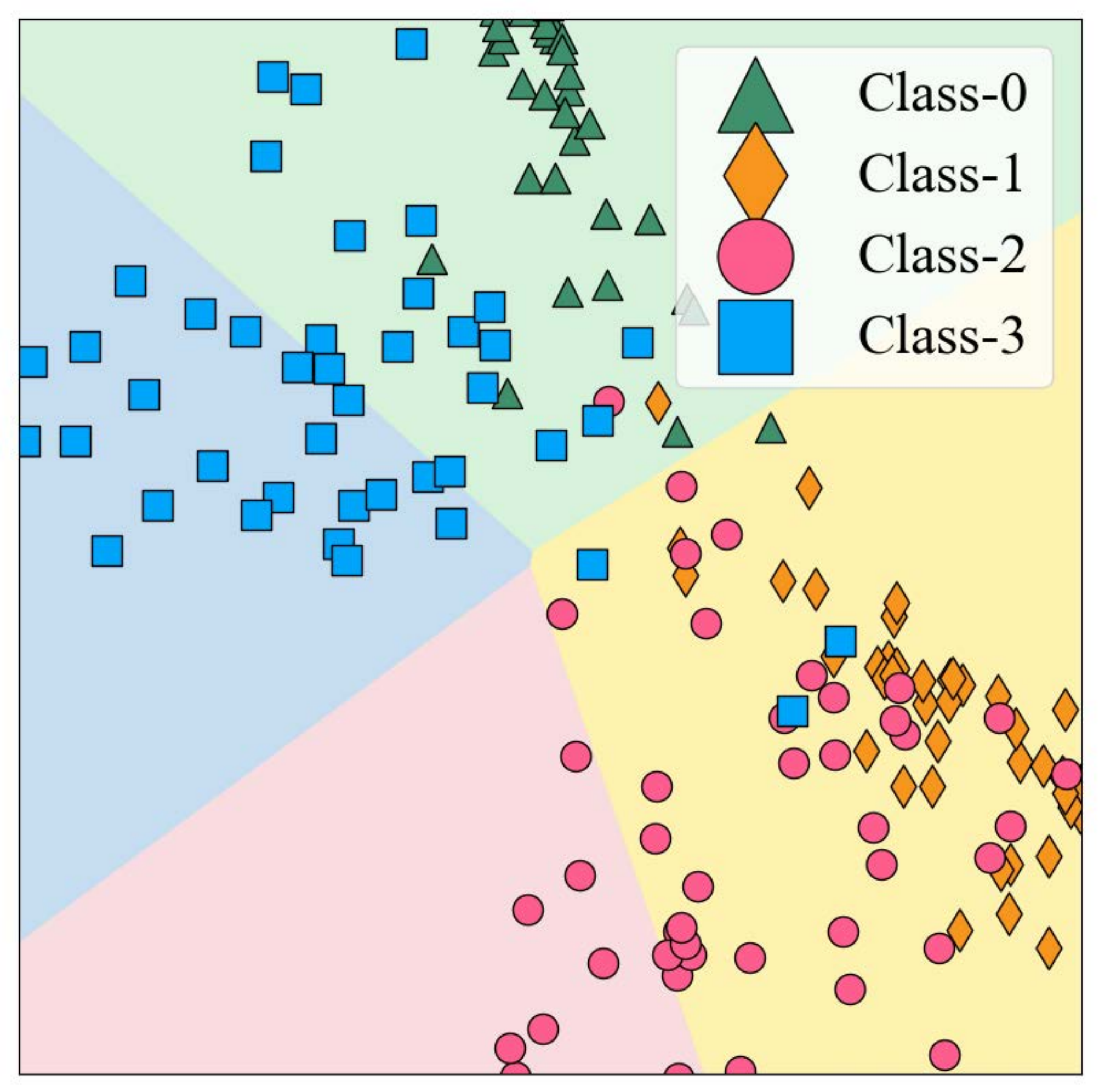}
	\end{minipage}
	\caption{\small \textbf{The training (top) and test (bottom) feature distributions along the training process, using ERM.} We study a four class imbalanced task, with different classes denoted by different colors/shapes ({\color{OliveGreen}Class-0}, {\color{Orange}Class-1} are major classes). There is a clear feature deviation (hence over-fitting) between training and testing.}
	\label{fig:ERM_toy}
	%\vspace{-5pt}
\end{figure}

\begin{figure}[t!]
\centering
	\begin{minipage}{0.24\linewidth}
		\centering
		\mbox{\small Epoch 20}
		\includegraphics[width=\linewidth]{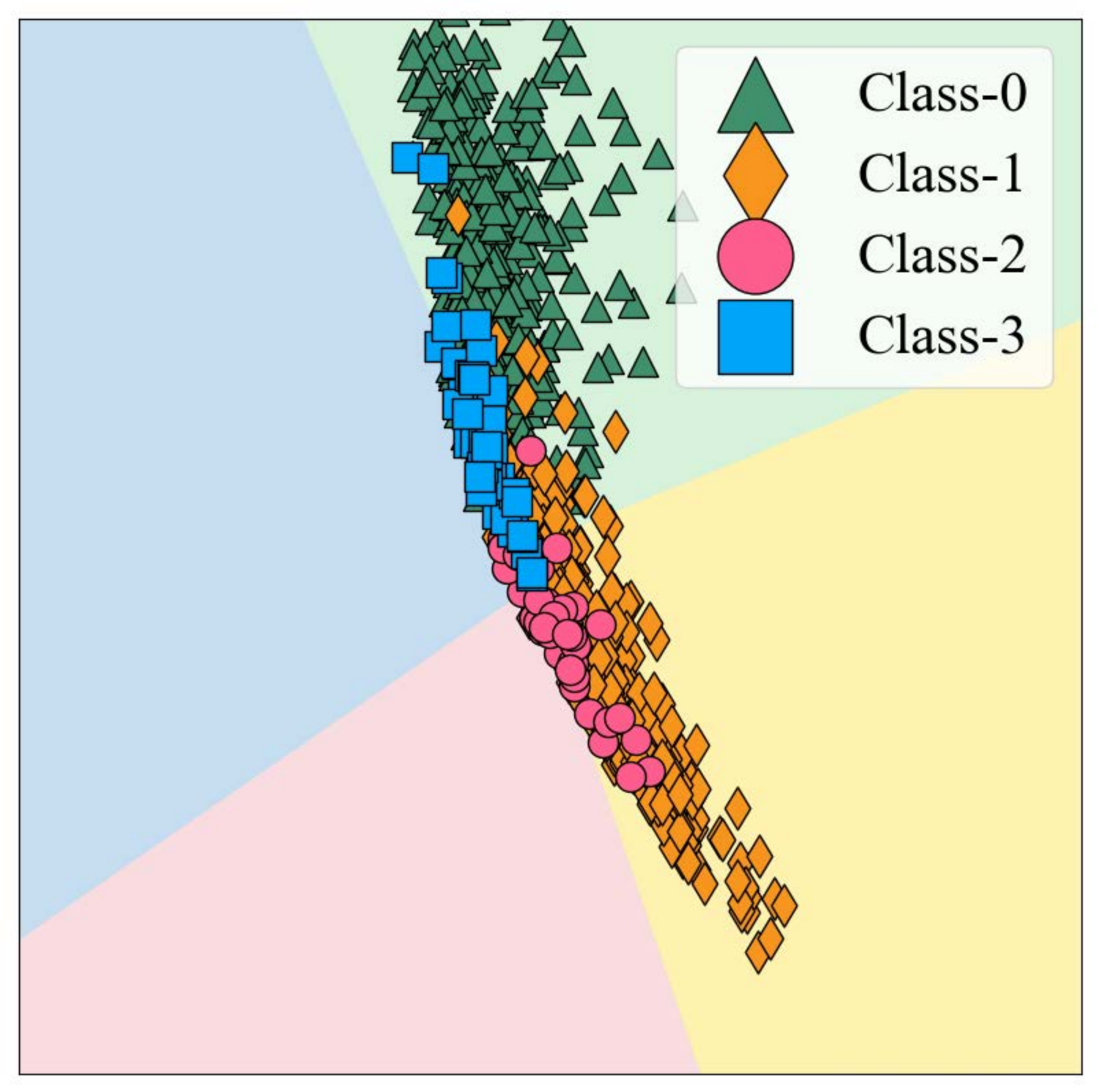}
	\end{minipage}
	\begin{minipage}{0.24\linewidth}
		\centering
		\mbox{\small Epoch 60}
		\includegraphics[width=\linewidth]{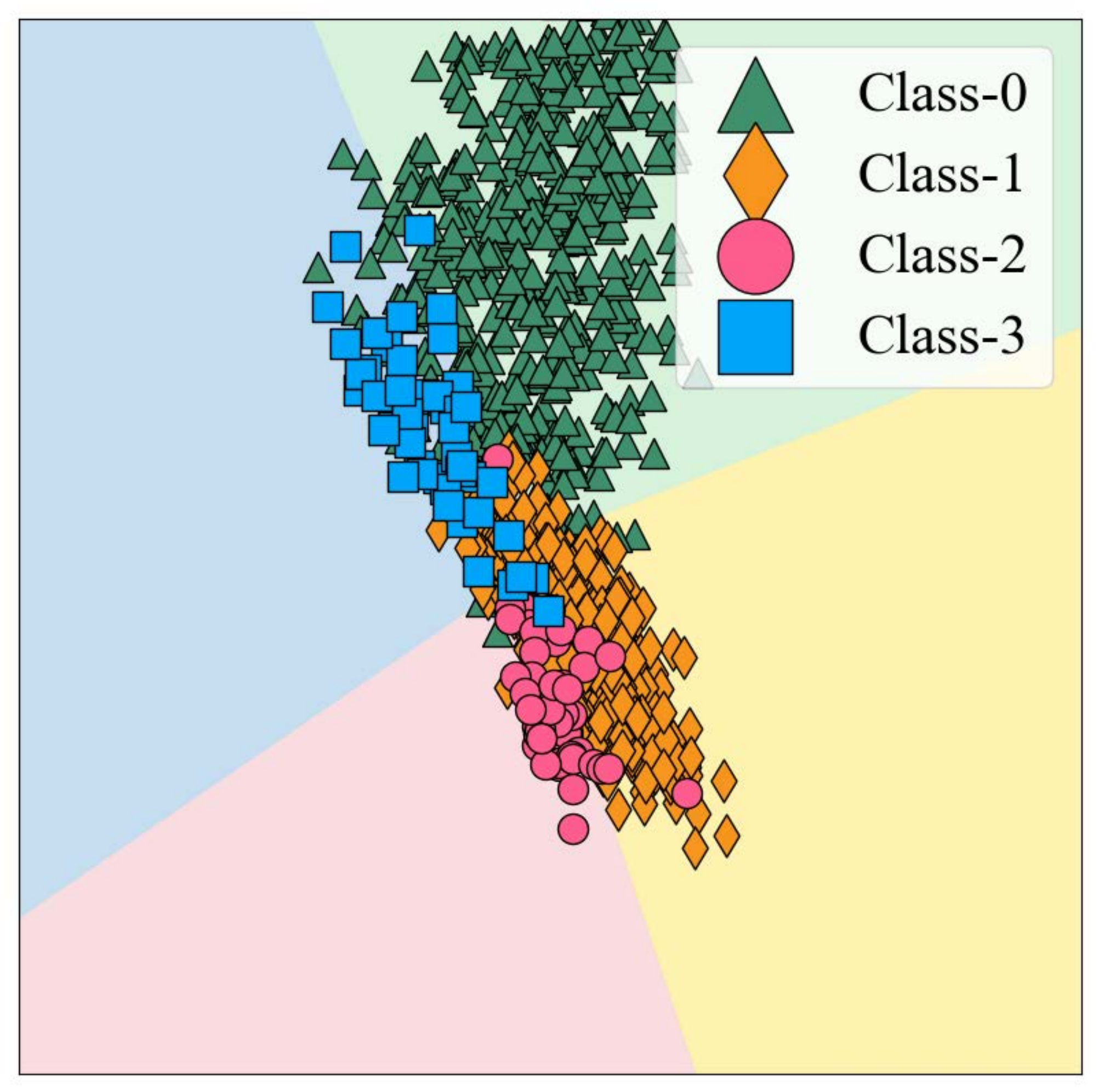}
	\end{minipage}
	\begin{minipage}{0.24\linewidth}
		\centering
		\mbox{\small Epoch 120}
		\includegraphics[width=\linewidth]{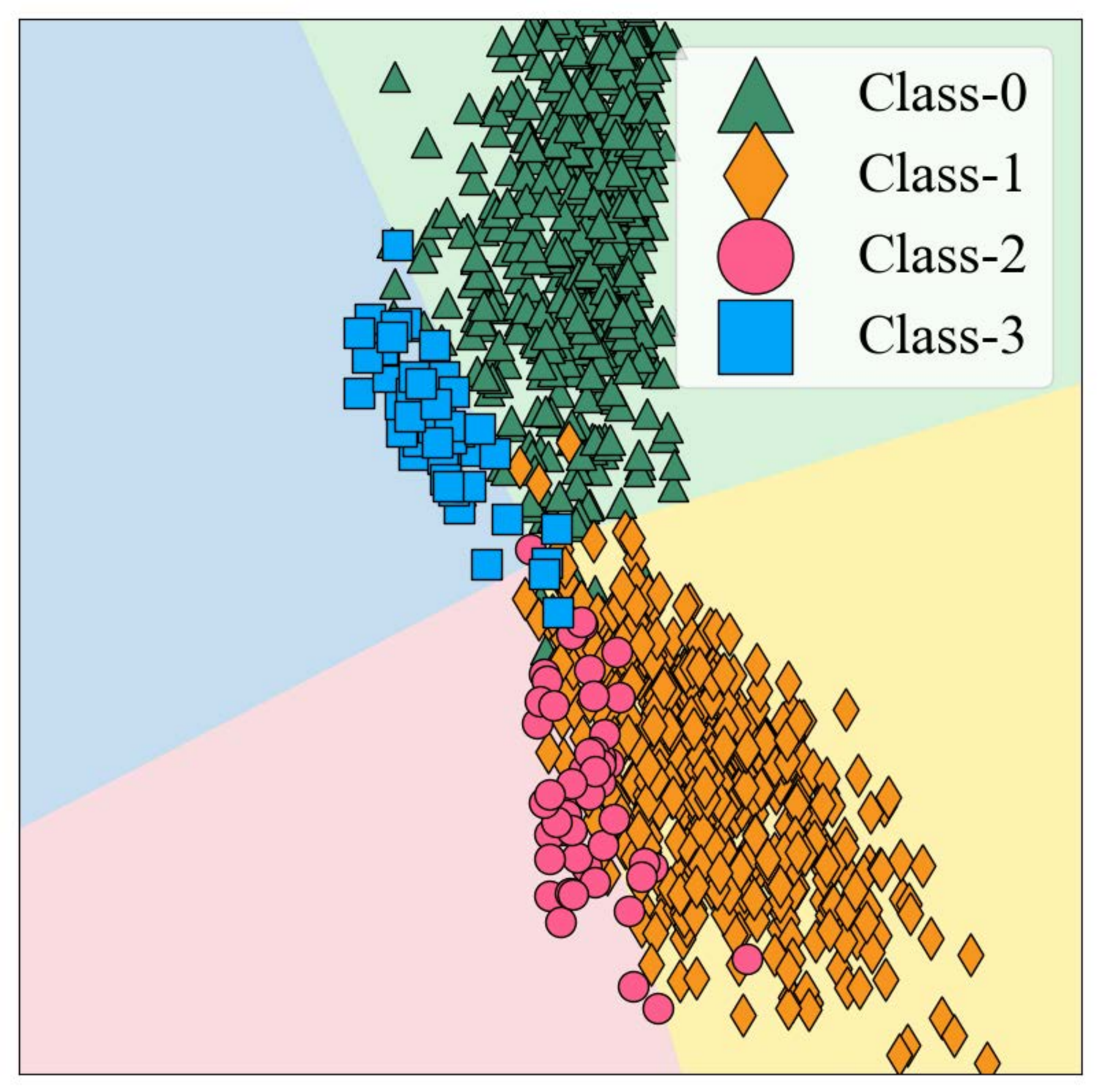}
	\end{minipage}
	\begin{minipage}{0.24\linewidth}
		\centering
		\mbox{\small Epoch 160}
		\includegraphics[width=\linewidth]{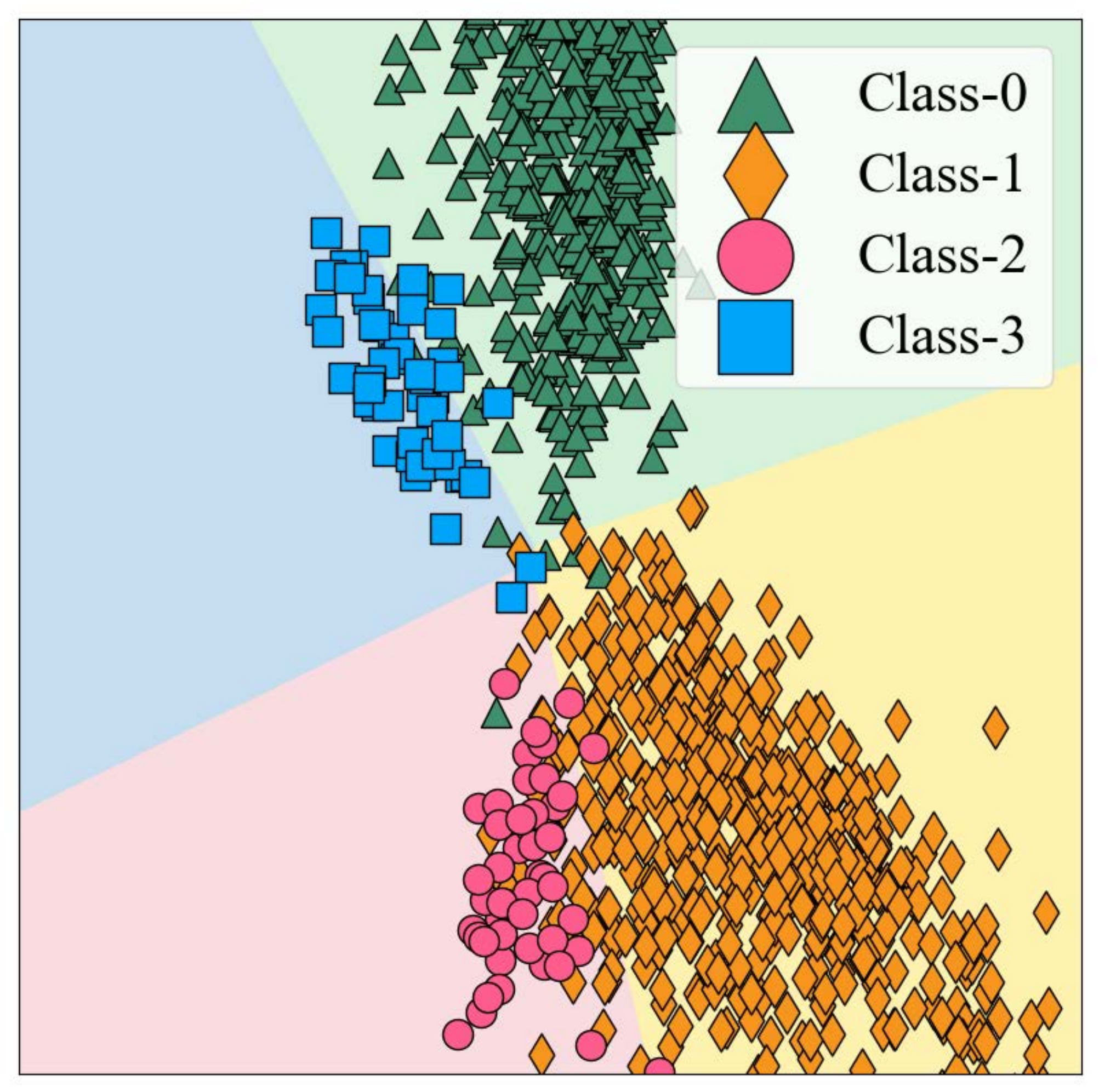}
	\end{minipage}\\
	\begin{minipage}{0.24\linewidth}
		\centering
		%\mbox{\small Epoch 20}
		\includegraphics[width=\linewidth]{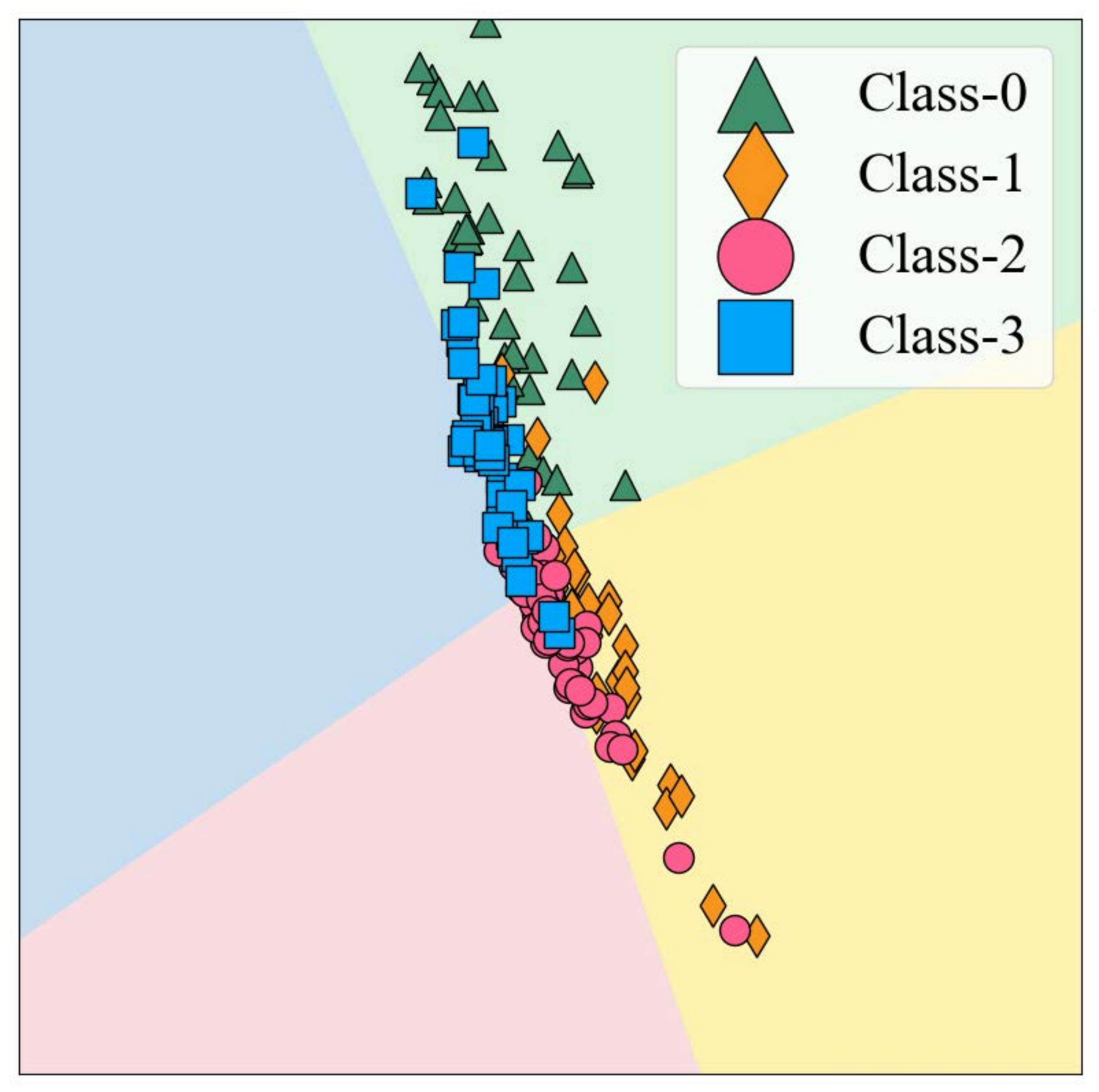}
	\end{minipage}
	\begin{minipage}{0.24\linewidth}
		\centering
		%\mbox{\small Epoch 60}
		\includegraphics[width=\linewidth]{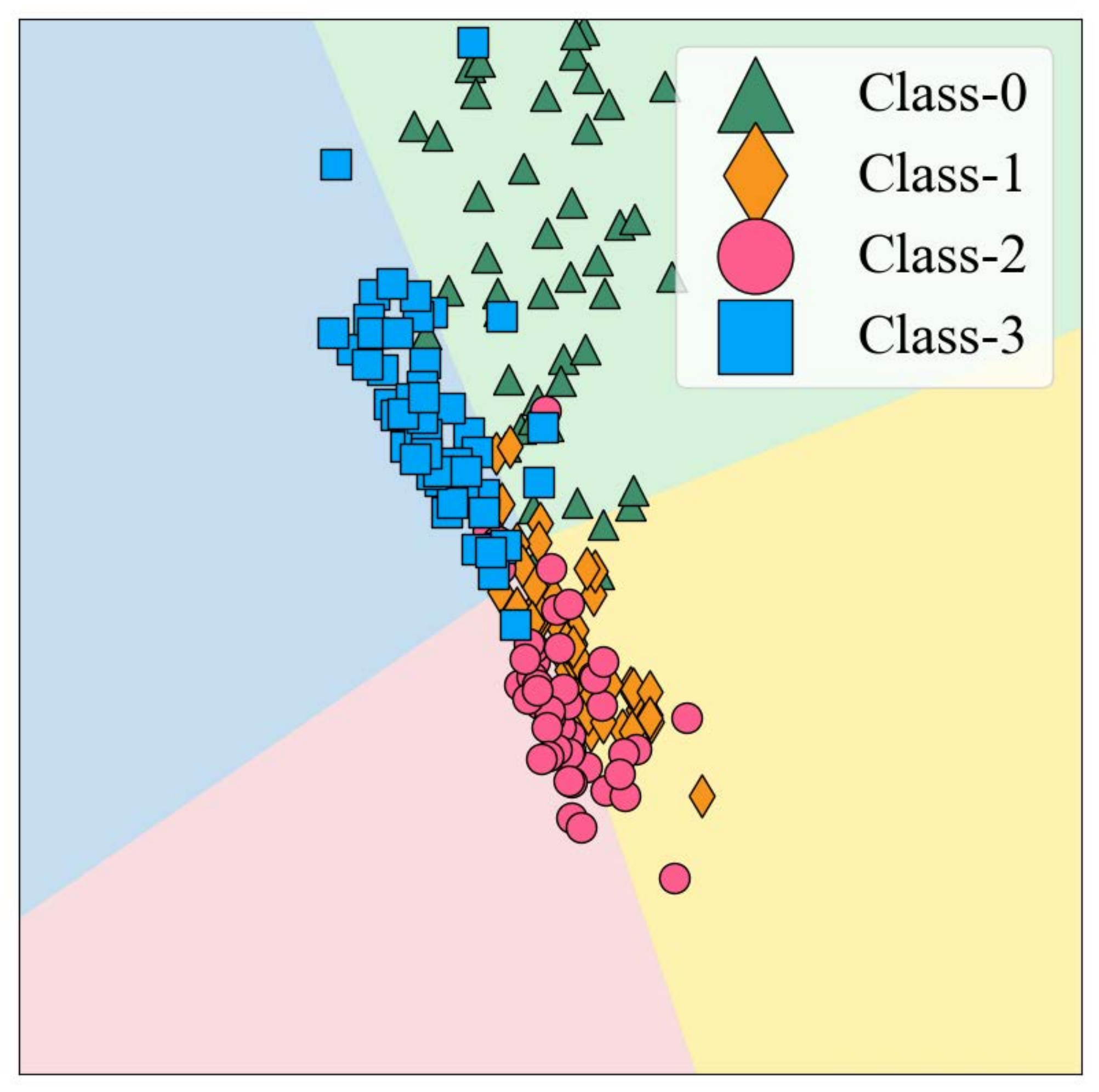}
	\end{minipage}
	\begin{minipage}{0.24\linewidth}
		\centering
		%\mbox{\small Epoch 120}
		\includegraphics[width=\linewidth]{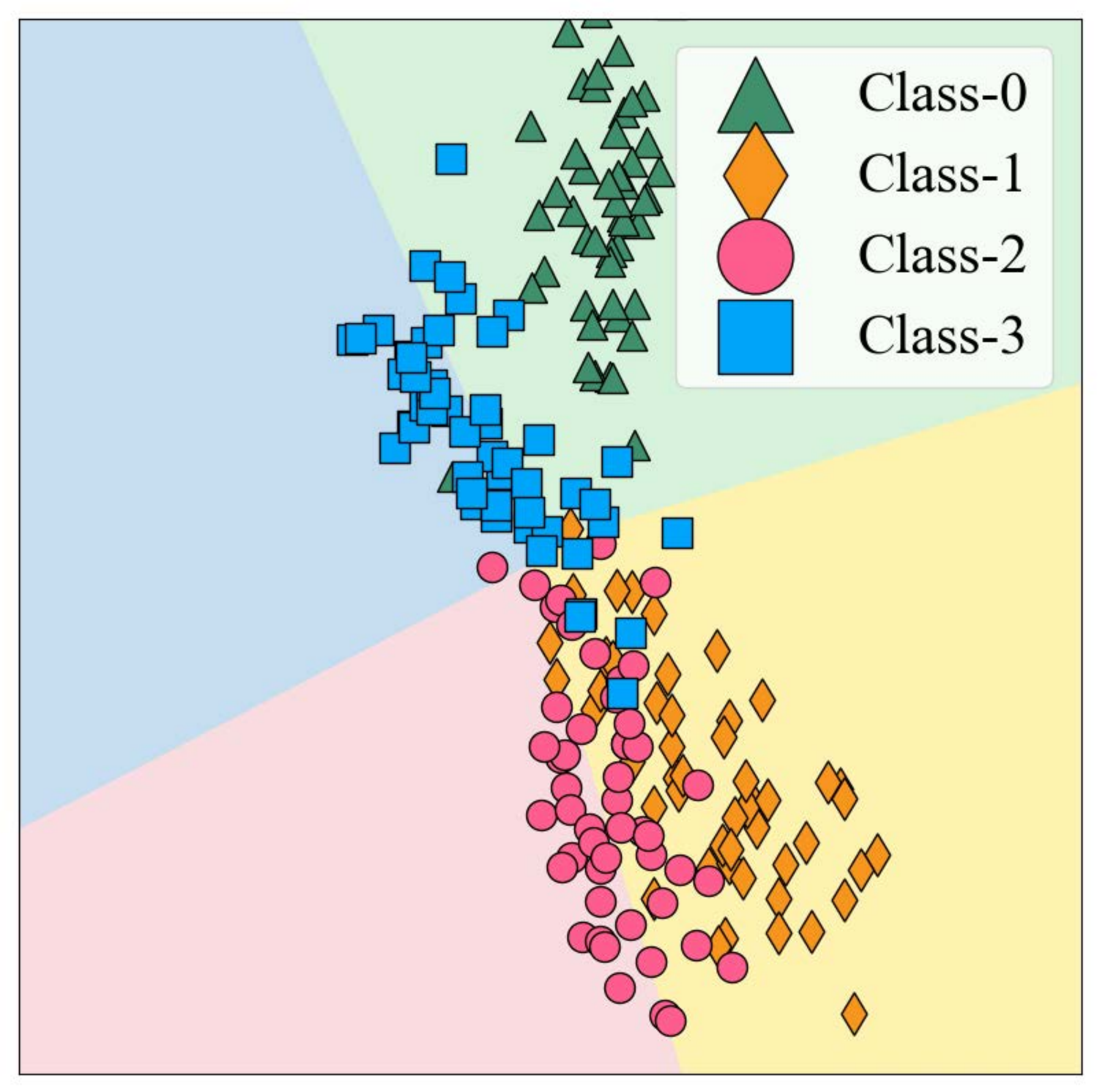}
	\end{minipage}
	\begin{minipage}{0.24\linewidth}
		\centering
		%\mbox{\small Epoch 160}
		\includegraphics[width=\linewidth]{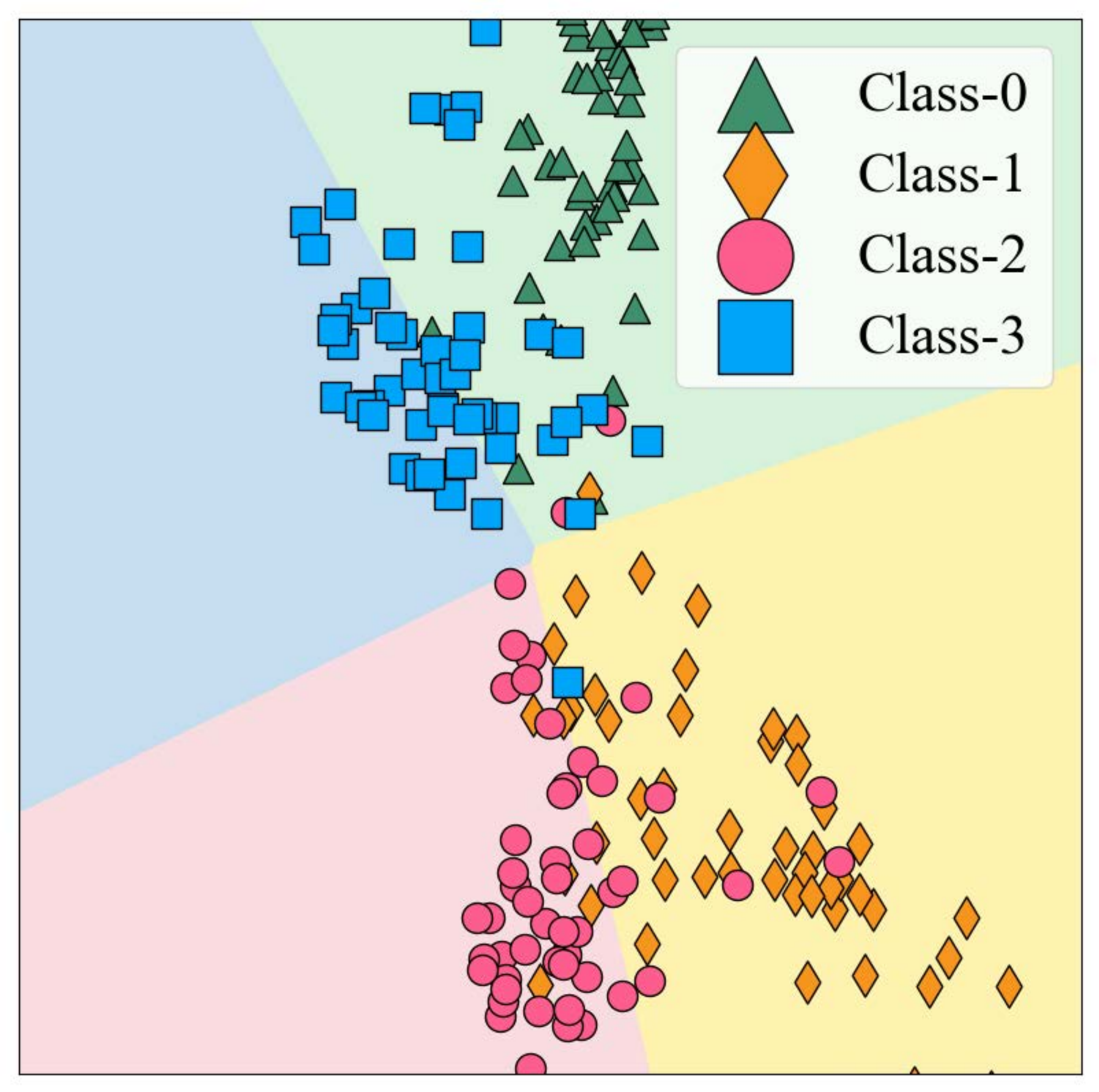}
	\end{minipage}
	\caption{\small \textbf{The training (top) and test (bottom) feature distributions along the training process, using \MFW.} We study a four class imbalanced task, with different classes denoted by different colors/shapes ({\color{OliveGreen}Class-0}, {\color{Orange}Class-1} are major classes). The feature deviation (over-fitting) between training and testing is reduced.}
	\label{fig:Mix_toy}
\end{figure}

To showcase the effect of \MFW, we conduct another experiment. We select four classes from CIFAR-10 \cite{krizhevsky2009learning}, and make their training data to be $5000, 5000, 50, 50$ per class: \ie, two major and two minor classes. The test data per class are $1,000$ samples. We use a ResNet-32 \cite{he2016deep} but add an additional linear projection layer to make the final feature dimension (\ie, right before the last fully-connected layer) to be $2$ for visualization. We then train the ResNet-32 with \MFW or with ERM, using the cross-entropy loss~(cf. \autoref{eq:CE}) for 200 epochs.
The initial learning rate is $0.1$ and it is decreased based on the cosine annealing rule. 

For \MFW, we mix the intermediate features after the second group of convolutional layers. We set $s = 0.5$ for major classes and $s=0$ for minor classes, and set $\alpha=2.0$ for the beta distribution.  In other words, the minor class features will not be weakened. After each training epoch, we plot the final two-dimensional features and the decision boundaries for both training and test data. Due to the page limit, only the results of 20, 60, 120, and 60 epochs are shown in \autoref{fig:ERM_toy} (for ERM) and \autoref{fig:Mix_toy} (for \MFW). We also sub-sample data in the figures to make it less crowded.

From \autoref{fig:ERM_toy} (for ERM), we see that at epoch 20, the training data (top row) of the minor classes are $100\%$ misclassified into the major classes. It thus leads to large gradients that try to push the minor class data away into their own territories. At the end of training, we see a nearly perfect separation for the training data. This is, however, not the case for the test data (bottom row). Specifically, at epoch 160, most of the test features of the minor classes are close to the boundaries between major and minor classes or even wrongly classified, essentially a case of \emph{feature deviations}.

Let us now look at \autoref{fig:Mix_toy} (for \MFW). There are four notable differences from \autoref{fig:ERM_toy}. First, at epoch 60, most of the minor class training data are correctly classified (or close to be); the major class features are kept close to the boundaries. The training progress is thus more equalized.
Second, with the \emph{gradient reduction} by \MFW, the competition between the major and minor classes are reduced. Even at epoch 160, the minor class training data are not overly pushed away from the boundaries. Third, the feature distributions of the training and testing data are much closer, indicating a smaller feature deviation. Finally, compared to \autoref{fig:ERM_toy}, a larger portion of the minor class test data are correctly classified at the end. As a result, the final model by \MFW outperforms that by ERM (83.65\% vs.  78.30\%).

%!TEX root=main.tex
\section{Experiment}
\label{s_exp}

\begin{table*}[t]
	\small
	%\vskip-5pt
	\centering
	\tabcolsep 3pt
	\caption{\small Test set accuracy (\%) on imbalanced CIFAR-10/-100. The best result of each setting (column) is in bold font.}
	\vspace{-2pt}
	\begin{tabular}{l|ccc|ccc||ccc|ccc}
		\addlinespace
		\toprule
		& \multicolumn{ 6}{c||}{CIFAR-10}                 & \multicolumn{ 6}{c}{CIFAR-100}                \\
		\midrule
		& \multicolumn{ 3}{c|}{Long-tailed} & \multicolumn{ 3}{c||}{Step} & \multicolumn{ 3}{c|}{Long-tailed} & \multicolumn{ 3}{c}{Step} \\
		Imbalance ratio $\rho$ & 200   & 100   & 10    & 200   & 100   & 10    & 200   & 100   & 10    & 200   & 100   & 10 \\
		\midrule
		ERM~\cite{cui2019class}   & 65.6  & 71.1  & 87.2  & 60.0    & 65.3  & 85.1  & 35.9  & 40.1  & 56.9  & 38.7  & 39.9  & 54.6 \\
		Focal~\cite{lin2017focal} & 65.3  & 70.4  & 86.8  & -     & 63.9  & 83.6  & 35.6  & 38.7  & 55.8  & -     & 38.6  & 53.5 \\
		CB~\cite{cui2019class}    & 68.9  & 74.6  & 87.5  & -     & 61.9  & 84.6  & 36.2  & 39.6  & 58.0    & -     & 33.8  & 53.1 \\
		LDAM-DRW~\cite{cao2019learning} & 74.6  & 77.0  & 88.2  & 73.6  & 76.9  & 87.8  & 39.5  & 42.0    & 58.7  & 42.4  & 45.4  & 59.5 \\
		$\tau$-Norm~\cite{kang2019decoupling} & 70.3  & 75.1  & 87.8  & 68.8  & 73.0    & 87.3  & 39.3  & 43.6  & 57.4  & {\bf 43.2} & 45.2  & 57.7 \\
		CDT~\cite{Ye2020CDT}   & 74.7  & 79.4  & 89.4 & 70.3  & 76.5  & 88.8  & 40.5 & 44.3  & 58.9 & 40.0    & 47.0    & 59.6 \\
		BBN~\cite{Zhou2020BBN}   &   -   & 79.8 & 88.3 &   -   &   -   &       &  -   & 42.6 & 59.1 &   -  &   -   & - \\
		M2M~\cite{Kim2020M2M}   &  -    & 79.1  & 87.5  &   -   &    -   &   -    &       & 43.5  & 57.6  &  -     &    -   & - \\
		Meta-Weight~\cite{shu2019meta} & 67.2  & 73.6 & 87.6 &   -    &    -   &       & 36.6 & 41.6 & 58.9 &   -    &   -    & - \\
		DA~\cite{Jamal2020Rethinking}    & 70.7 & 76.4 & 88.9 & -      &   -    &    -   & 39.3 & 43.4 & 59.6 &     -  &    -   & - \\
		Remix-DRW~\cite{chou2020remix} &   -    & 79.8 & 89.0 &    -   & 77.9 & 88.3 &   -    & {\bf 46.8} & {\bf 61.2} & -      & 46.8 & 60.4 \\
		De-confound-TDE~\cite{tang2020long} &   -    & {\bf 80.6} & 88.5 &    -   & - & - &   -    & 44.1 & 59.6 & -      & - & - \\
		\midrule
		\MFW & 73.2 & 78.5 & {\bf 89.8} & 75.4 & 80.1 & 89.6 & 40.7 & 44.7 & 60.1 & 42.5 & 46.9  & 61.2 \\
		\MFW w/ DRW & {\bf 75.0} & 79.8 & 89.7 & {\bf 78.8} & {\bf 81.6} & {\bf 89.9} & {\bf 41.4} & 46.0 & 59.1 & 43.0 & {\bf 48.4} & {\bf 61.6} \\
		\bottomrule
	\end{tabular}
	\label{tab:cifar-table}
	%\vskip-5pt
\end{table*}

\subsection{Setup}
\label{sec:setup}
\noindent\textbf{Datasets.} We validate \MFW on five datasets. \textbf{CIFAR-10} and \textbf{CIFAR-100} \cite{krizhevsky2009learning} are for image classification with 32 $\times$ 32 images. There are $50,000$ training and $10,000$ test images from 10 and 100 classes, respectively. \textbf{Tiny-ImageNet}~\cite{le2015tiny} has 200 classes. Each class has 500 training and 50 validation images of $64\times 64$ pixels.
\textbf{iNaturalist}~\cite{van2018inaturalist} (2018 version) is a natural large-scale long-tailed dataset containing 437,513 training images from 8,142 classes and there are 3 validation images per class. The image resolution is $224\times 224$.
\textbf{Tiered-ImageNet}~\cite{Ren2018Meta} is a subset of ImageNet~\cite{deng2009imagenet} widely used in few-shot learning. The image resolution is $84\times 84$. We use {Tiered-ImageNet} to synthesize a large-scale step imbalanced dataset. We treat the $351$ many-shot classes as the major classes, each with around $1,000$ training images, and the $160$ few-shot classes as the minor classes, each with $5$ training instances. All the classes have $50$ test instances. See the supplementary for details.

\begin{table*}[t]
	\small
	%\vskip-5pt
	\centering
	\caption{\small Top-1/-5 validation set accuracy (\%) on imbalanced Tiny-ImageNet. The best result of each setting (column) is in bold font.}
	\vspace{-2pt}
	\begin{tabular}{l|cccc|cccc}
		\addlinespace
		\toprule
		& \multicolumn{ 4}{c|}{Long-tailed}      & \multicolumn{ 4}{c}{Step}     \\
		\midrule
		Imbalance ratio $\rho$ & 100 Top-1 & 100 Top-5 & 10 Top-1 & 10 Top-5 & 100 Top-1 & 100 Top-5 & 10 Top-1 & 10 Top-5 \\
		\midrule
		ERM~\cite{cao2019learning}   & 33.8  & 57.4  & 49.7  & 73.3  & 36.2  & 55.9  & 49.1  & 72.9 \\
		CB~\cite{cui2018large}    & 27.3  & 47.4  & 48.4  & 71.1  & 25.1  & 40.9  & 45.5  & 66.8 \\
		LDAM-DRW~\cite{cao2019learning} & 37.5  & 60.9  & \bf 52.8  & \bf 76.2  & 39.4  & 61.9  & 52.6  & \bf 76.7 \\
		$\tau$-Norm~\cite{kang2019decoupling} & 36.4  & 59.8  & 49.6  & 72.8  & 40.0    & 61.9  & 51.7  & 75.2 \\
		CDT~\cite{Ye2020CDT}   & \bf 37.9  & \bf 61.4  & 52.7  & 75.6  & 39.6  & 61.5  & 53.3  & 76.2 \\
		\midrule
		\MFW & 35.4  & 59.2  & 51.0    & 73.4  & \bf 40.4 & \bf 62.9  & 52.9 & 76.3 \\
		\MFW w/ DRW & 36.2  & 59.8  & \bf 52.8 &  74.5  & 40.0 & 61.2 & \bf 54.3 & \bf 76.7 \\
		\bottomrule
	\end{tabular}\label{tab:tiny-imagenet-table}
	\vskip-10pt
\end{table*}

\noindent\textbf{Setup.} 
We follow~\cite{cao2019learning,cui2019class} to create \emph{imbalanced} CIFAR-10, CIFAR-100, and Tiny-ImageNet with different imbalance ratios $\rho = N_\text{max}/N_\text{min}\in\{10, 100, 200\}$. Two types of imbalanced data are investigated, \ie, the \emph{long-tailed} (LT) imbalance where the number of training instances exponentially decayed per class, and the \emph{step} imbalance where the size of the second half of classes are proportional to a fixed ratio w.r.t. the first half head classes.
The test and validation sets remain unchanged and balanced. We re-index classes so that smaller indices have more training instances. 

We follow existing works \cite{cao2019learning,cui2019class,Zhou2020BBN,Lee2019Meta} to use ResNet \cite{he2016deep}: ResNet-32 for CIFAR, ResNet-18 for Tiny-ImageNet, ResNet-12 for Tiered-ImageNet, and ResNet-50 for iNaturalist. 
We adapt the code from \cite{cao2019learning,cui2019class} for CIFAR, from \cite{Lee2019Meta} for Tiered-ImageNet, and from \cite{Zhou2020BBN} for iNaturalist. Please see the supplementary material for details.

Following~\cite{cui2019class,cao2019learning,Zhou2020BBN}, the test set accuracy (validation set accuracy on Tiny-ImageNet and iNaturalist) are reported for evaluation. See the supplementary material for details.

\subsection{Implementation of \MFW}
\label{ssec:mfw_imp}
We apply the cross-entropy loss in \autoref{eq:CE} to train ResNet with \MFW. Considering a residual block and the convolutional layers before the first residual block each as a group of convolutional layers, we apply \MFW after the second groups of convolutions, unless stated otherwise.

We design the weight function $s$ in Algorithm~\ref{a_MFW} as follows. First, the weight should be monotonically increasing from minor to major classes. Second, major and minor classes get weights around $0.5$ (more weakening) and $0.0$ (no weakening), respectively.
To take the number of instances per class $N_c$ into account, we define $s$ as follows
\begin{align}
s(N_c) = 0.5\times \sigma(\frac{N_c - \mu}{\beta \cdot \gamma})\label{eq:lambda_n},
\end{align}
where $\sigma(a) = \frac{1}{1+\exp(-a)}$ is the sigmoid function, which is widely used to squash real values into the range $[0,1]$. $\mu$ and $\gamma$ are the geometric mean and standard deviation over $\{N_c\}_{c=1}^C$, which normalize the weights with respect to all classes\footnote{Subtracting the mean and dividing by the standard deviation are common practices to normalize function inputs (\eg, z-score).}. We use the geometric mean because it is less sensitive to extremely large $N_c$ than the arithmetic mean, making the weight stable in both long-tailed and step settings. 
The scale $\beta$ controls the softness of weights. We set $\beta=2.0$ for the long-tailed case and $\beta=0.01$ for the step case. 
We tune the beta distribution coefficient $\alpha$ on a small held-out set from the training data. Details are in the supplementary.

Inspired by \cite{cao2019learning,chou2020remix}, we also apply Deferred Re-Weight (DRW) after training 80\% of epochs. DRW applies an instance-specific weights over the loss function, which stresses the optimization for minor classes. We follows the same weighting strategy as \cite{cao2019learning}. 

\begin{figure*}[t!]
	\begin{minipage}{0.245\linewidth}
		\centering
		\mbox{\small Long-tailed (ERM)}
		\includegraphics[width=\linewidth]{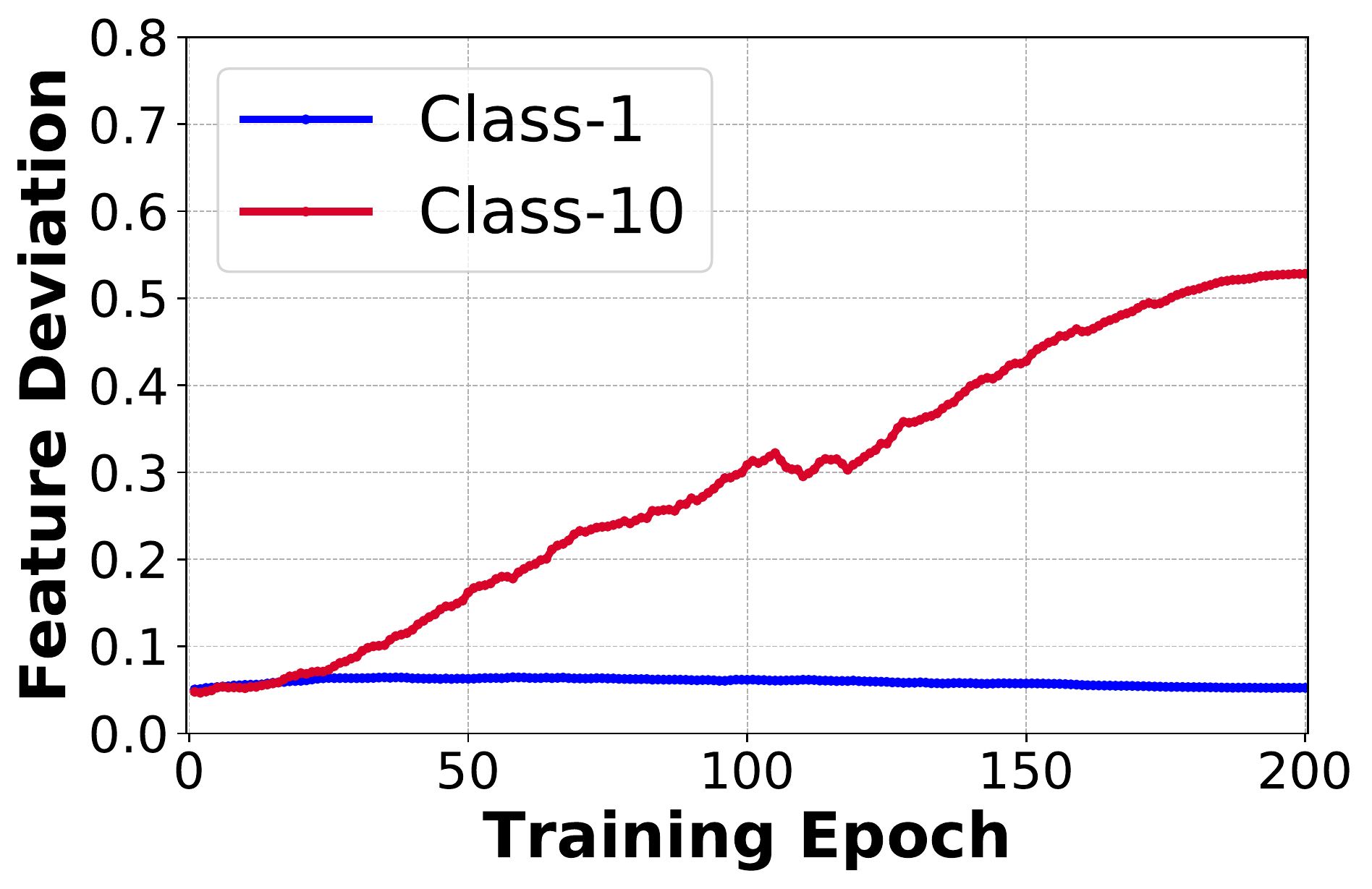}
	\end{minipage}
	\begin{minipage}{0.245\linewidth}
		\centering
		\mbox{\small Long-tailed (\MFW)}
		\includegraphics[width=\linewidth]{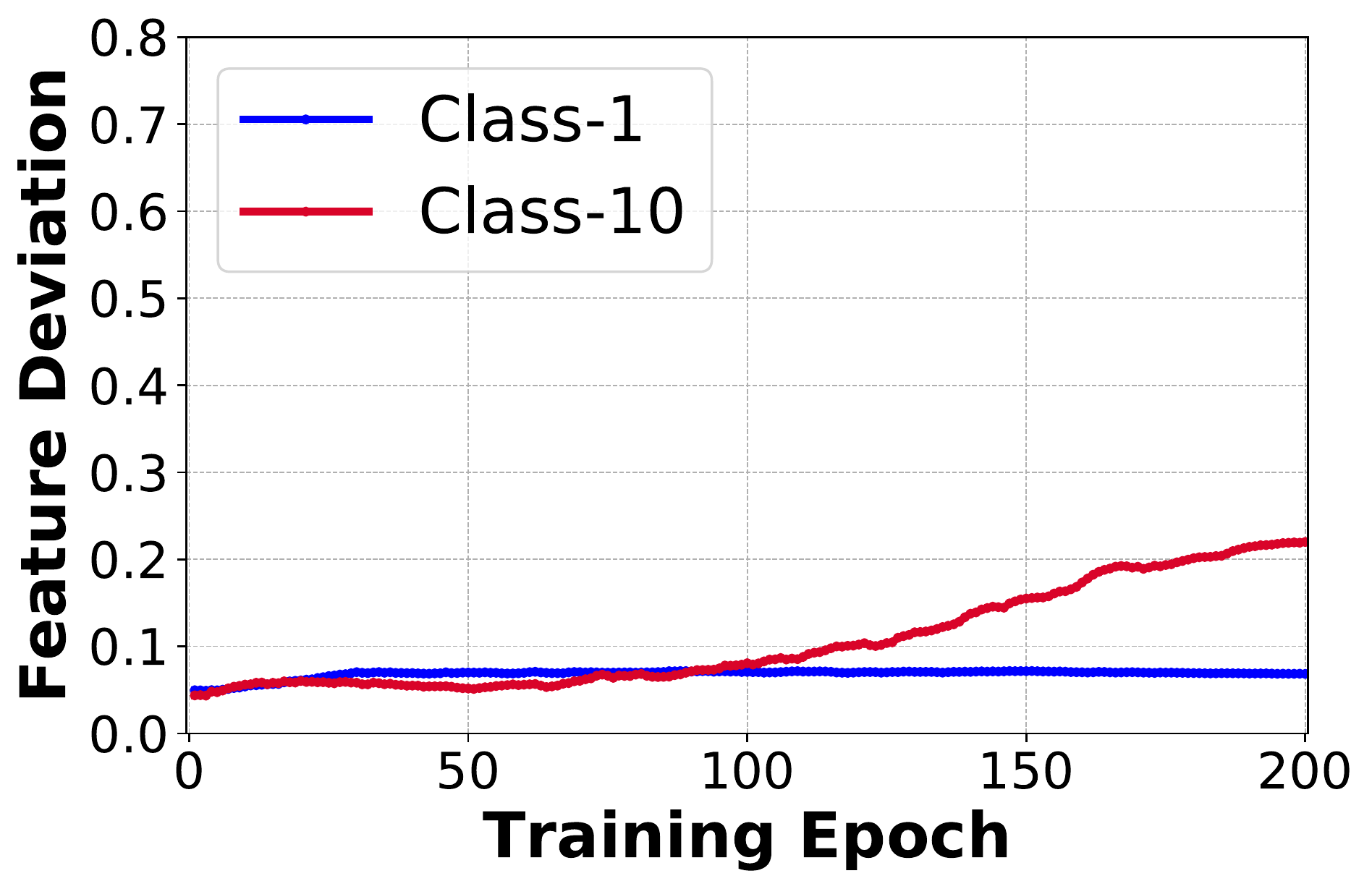}
	\end{minipage}
	\begin{minipage}{0.245\linewidth}
		\centering
		\mbox{\small Step (ERM)}
		\includegraphics[width=\linewidth]{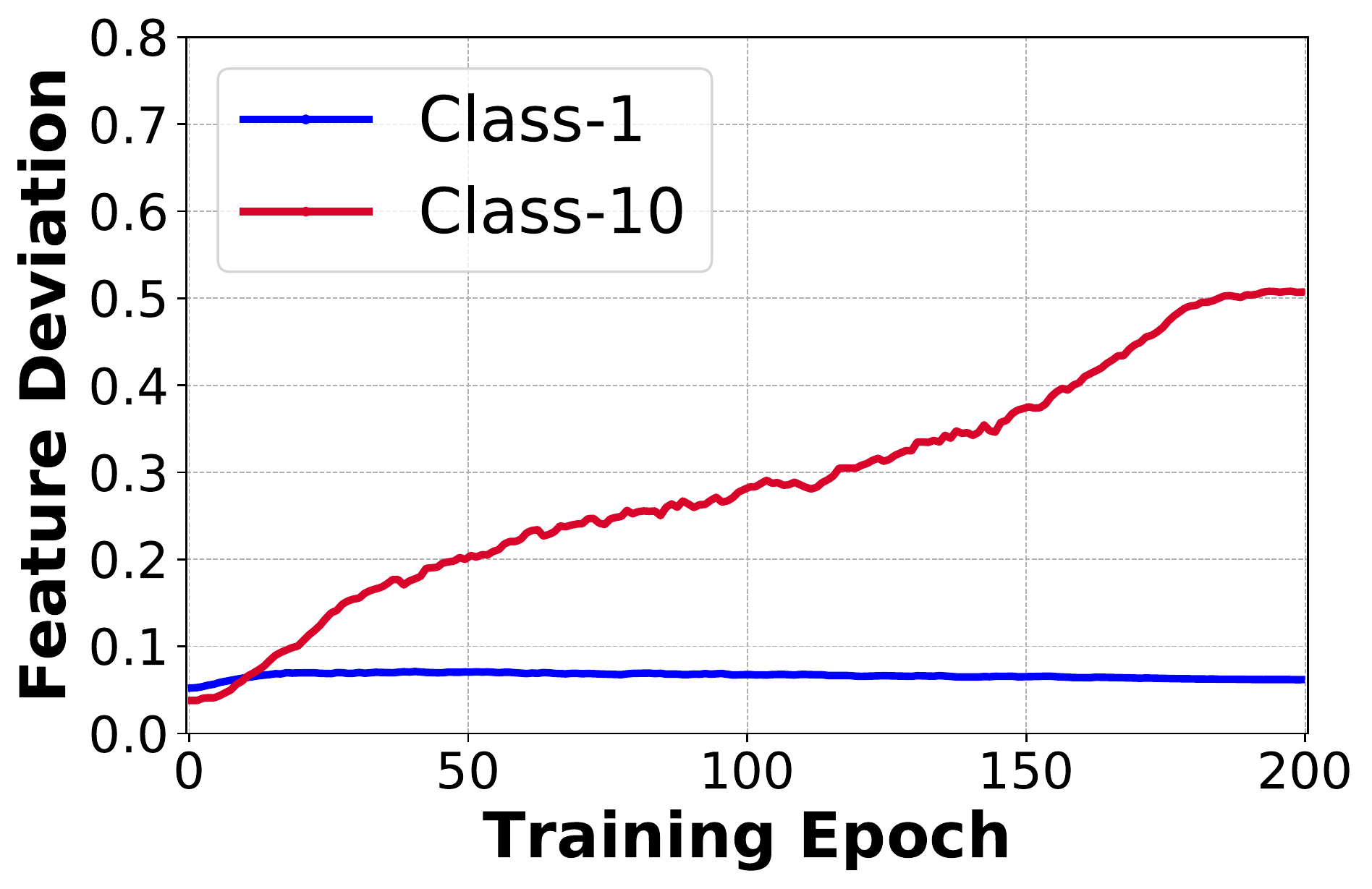}
	\end{minipage}
	\begin{minipage}{0.245\linewidth}
		\centering
		\mbox{\small Step (\MFW)}
		\includegraphics[width=\linewidth]{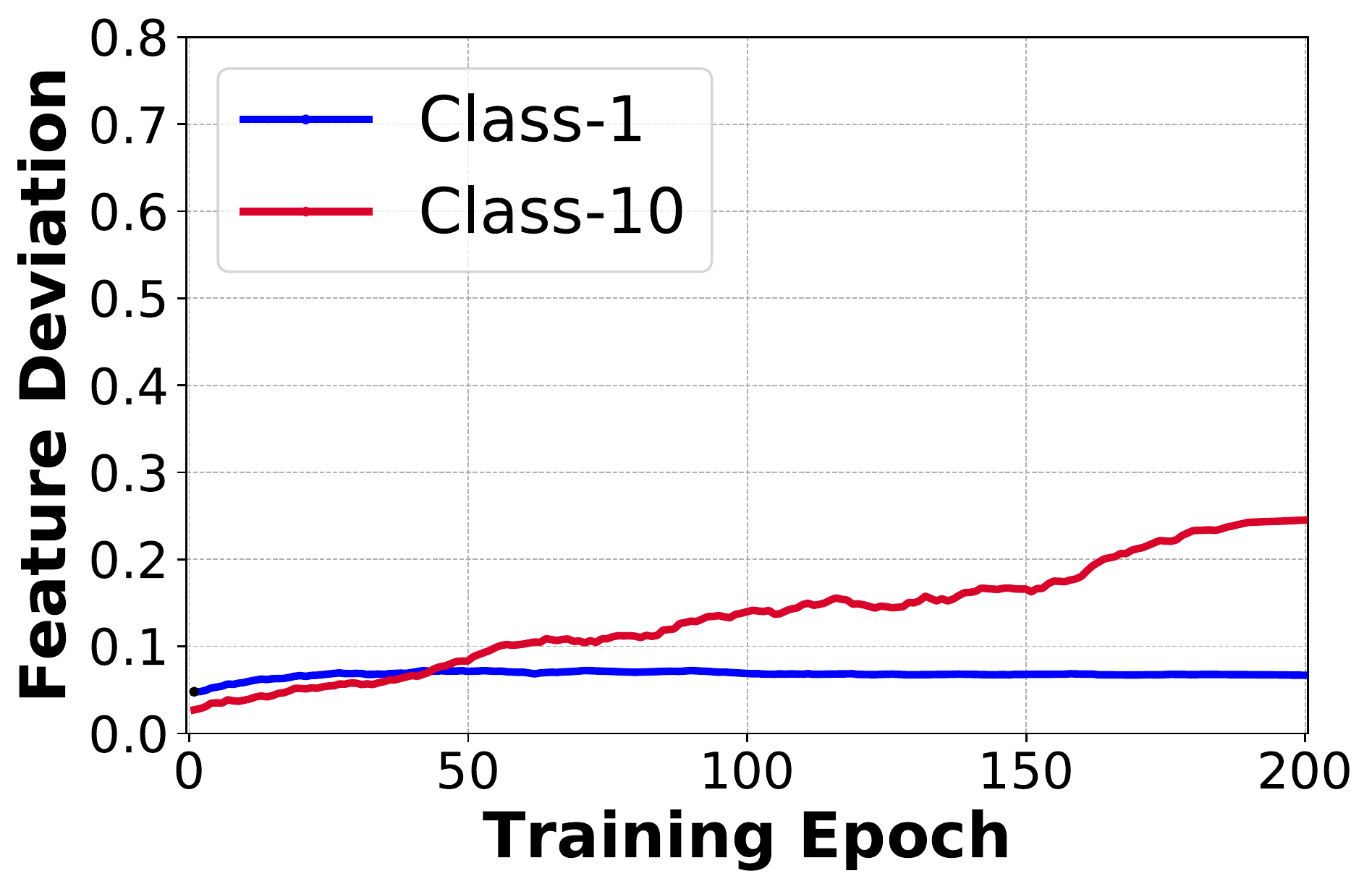}
	\end{minipage}
	\caption{\small \textbf{Feature deviation between the training and test data per class along the training progress.} We experiment on CIFAR-10, using both the long-tailed and the step settings ($\rho=100$).
	We only showed the most major ($c = 1$, with $5,000$ training samples) and minor classes ($c = 10$, with $50$ training samples) for clarity.
	The formulation of deviation is in~\autoref{ss_abs}. As the number of training epochs increases, the deviation increases, while \MFW can achieve a much smaller deviation.}
	\label{fig:deviation}
	\vspace{-10pt}
\end{figure*}

\subsection{Results}
\label{sec:result}

\noindent\textbf{CIFAR.}
We extensively examine CIFAR-10 and CIFAR-100 with imbalance ratios $\rho\in\{10, 100, 200\}$ on both long-tailed and step cases. Results are shown in \autoref{tab:cifar-table}. 
\MFW (without DRW) is on a par with or better than the compared approaches. With DRW, \MFW achieves the best performance in most of the settings. 

\emph{It is worth noting that \MFW obtains particularly high accuracy on the \textbf{step imbalance scenario} that most of the other methods struggled.} In contrast to the long-tailed setting where there is a smooth transition of class sizes between the major and minor classes, the step setting only has two extremes (\ie, major or minor classes), whose sizes are different by a ratio $\rho$. We attribute the superior performance of \MFW
to its inner working, which aims to resolve the \emph{over-competition between the major and minor classes}. Such a competition is stronger in the step case than in the long-tailed case, and that is why \MFW excels. 

\begin{table}[t]
    \small
	\centering
	\caption{\small { Top-1/-5 test accuracy (\%) on Step Tiered-ImageNet.}}
	\vskip-2pt
    \begin{tabular}{l|cc}
    \addlinespace
    \toprule
          & Top1  & Top5 \\
    \midrule
    ERM   &  41.7 & 58.1 \\
    LDAM~\cite{cao2019learning}  & 42.3 & 63.1 \\
    CDT~\cite{Ye2020CDT}   & 43.8 &  64.2 \\
    cRT~\cite{kang2019decoupling}  & 44.4 & 66.5 \\
    $\tau$-norm~\cite{kang2019decoupling} &  43.0 & 62.4 \\
    LWS~\cite{kang2019decoupling} & 42.6 & 58.8 \\
    Remix~\cite{chou2020remix} & 43.4 & 62.6 \\
	\midrule
    \MFW &  46.1   &  67.5 \\
    \MFW w/ DRW &  \textbf{46.4}  & \textbf{67.8} \\
    \bottomrule
    \end{tabular}
	\label{tab:tiered-table}
	%\vskip-5pt
\end{table}

\noindent\textbf{Tiny-ImageNet. (\autoref{tab:tiny-imagenet-table})} Our \MFW again performs particularly well in the challenging step setting.

\noindent\textbf{Step Tiered-ImageNet. (\autoref{tab:tiered-table})}
Tiered-ImageNet is a large-scale step imbalanced dataset ($511$ classes) whose imbalanced ratio $\rho > 200$. We re-implement all the baselines. Our \MFW outperforms  all methods, which demonstrate the strong ability of \MFW on step imbalanced cases.

\noindent\textbf{iNaturalist. (\autoref{tab:inat-table})}
iNaturalist has 8,142 classes and many of them have scarce instances, making it particularly challenging. Our \MFW outperforms nearly all but the Remix approach (worse by $\sim1\%$).  As large-scale training is sensitive to batch size, the fact that Remix uses a batch size of 256 while we use 128 (due to the computational constraint) might contribute to the difference.

\begin{table}[t]
    \small
	\centering
	\caption{\small { Top-1/-5 validation accuracy (\%) on iNaturalist}. We present results by training with 90/180 epochs in the form ``A/B''.}
	\vskip-2pt
    \begin{tabular}{l|ll}
    \addlinespace
    \toprule
          & \hspace{10pt} Top1  & \hspace{10pt} Top5 \\
    \midrule
    ERM   & 58.8 / 64.3  & 80.1 / 84.5 \\
    CB~\cite{cui2019class}    & 61.5 / \hspace{5pt} - \hspace{5pt} & 80.9 / \hspace{5pt} - \hspace{5pt}\\
    LDAM~\cite{cao2019learning}  & 64.6 / 66.1  & 83.5 / \hspace{5pt} - \hspace{5pt}\\
    LDAM-DRW~\cite{cao2019learning} & 68.0 / 68.6    & 85.2 / 85.3 \\
    CDT~\cite{Ye2020CDT}   & 63.7 / 69.5  & 82.5 / 86.8 \\
    BBN~\cite{Zhou2020BBN}   & 66.3 / 69.6 & \hspace{10pt} - \\
    cRT~\cite{kang2019decoupling}   & 65.2 / 67.6  & \hspace{10pt} - \\
    $\tau$-norm~\cite{kang2019decoupling} & 65.6 / 69.3  & \hspace{10pt} - \\
    Remix-DRW~\cite{chou2020remix} & \hspace{5pt} - \hspace{5pt}/ 70.5 & \hspace{5pt} - \hspace{5pt}/ 87.3 \\
	\midrule
    \MFW &   65.5 / 67.3  &  85.3 / 85.8 \\
    \MFW w/ DRW &  66.7 / 69.6  & 85.5 / 86.1 \\
    \bottomrule
    \end{tabular}
	\label{tab:inat-table}
	%\vskip-5pt
\end{table}

\subsection{Ablation Studies}
\label{ss_abs}
We conduct further analysis on CIFAR-10 (for $\rho=100$).

\noindent\textbf{Can \MFW reduce the feature deviation?}
We follow~\cite{Ye2020CDT} to compute the feature deviation.
We extract $\ell_2$-normalized features, compute feature means in training and test data for each class $c$, and calculate their Euclidean distance $dis(c) = $
\begin{align}
    & \frac{1}{R}\sum_{r=1}^R \|\text{mean}(S_K(\{f_{\vtheta}(\vx_{\text{train}}^{(c)})\}))-\text{mean}(\{f_{\vtheta}(\vx_{\text{test}}^{(c)})\})\|_2. \nonumber
\end{align}
$S_K$ is a subsampling (of $K$ examples) over training samples per class before computing the mean, which is to alleviate the estimation variance resulting from different class sizes. We follow \cite{Ye2020CDT} to perform $R=1,000$ subsampling rounds and set $K$ as the minor class size. The larger the $dis(c)$ is, the larger the feature deviation is, which implies severer over-fitting (thus worse accuracy).
\autoref{fig:deviation} shows the result. \MFW notably reduces the feature deviation, justifying our claims that feature deviation is likely caused by the exaggerated feature gradients on the minor classes. 

\noindent\textbf{Which layer for \MFW? (\autoref{tab:cifar-abltaion})}
We apply \MFW after different groups of convolutions. Intermediate layers yield the highest accuracy. All of them outperform ERM.

\noindent\textbf{Label mixing.} We compare \MFW to mixup \cite{zhang2018mixup} and manifold mixup \cite{verma2019manifold}. Details are in the supplementary.

\begin{table}[t]
	\small
	\centering
	\tabcolsep 3pt
	\caption{\small Test accuracy (\%) on CIFAR-10/-100. We apply \MFW after each convolutional group and compare to \cite{zhang2018mixup,verma2019manifold}.}
	\vskip-2pt
	\begin{tabular}{l|cc||cc}
		\addlinespace
		\toprule
		& \multicolumn{ 2}{c||}{CIFAR-10}                 & \multicolumn{ 2}{c}{CIFAR-100}                \\
		\midrule
		Imbalance ratio $\rho=100$ & Long-tailed & Step & Long-tailed & Step \\
		\midrule
		ERM & 71.1 & 65.8 & 40.1 & 39.9 \\
		\midrule
		\MFW (Input layer)  & 76.3 & 68.1 & 41.2 & 43.1 \\
		\MFW (1st Group) & 77.3 & 72.4 & 42.3 & 44.3 \\
		{\MFW (2nd Group)} & \textbf{78.5} & \textbf{80.1}  & \textbf{44.7} & \textbf{46.9} \\
		\MFW (3rd Group) & 77.1 & 69.8 & 43.1 &  46.3 \\
		\MFW (4th Group)  & 75.0 & 67.4 & 40.3 & 42.2\\
		\bottomrule
	\end{tabular}
	\label{tab:cifar-abltaion}
	\vskip-5pt
\end{table}

%!TEX root=main.tex
\section{Conclusion}
Class-imbalanced deep learning is a fundamental problem and a practical issue to resolve in computer vision. 
In this paper, we take a new perspective to tackle it, which is to investigate how imbalanced data affect the training progress of a neural network. Namely, how major classes and minor classes are fitted along the epochs. We found a huge discrepancy: a network tends to fit the major classes first and then the minor ones, resulting in large gradients for the minor class data that may eventually lead to feature deviation and over-fitting. We propose a fairly simple yet mathematically sound approach \MFW to effectively balance the training progress and reduce the minor class gradients. \MFW performs well on multiple benchmark datasets, especially on the challenging step imbalanced scenario.
\label{s_disc}

%\section*{Acknowledgement}
{\small \noindent\textbf{Acknowledgement.} The research was supported by National Key R\&D Program of China (2020AAA0109401) and NSFC (61773198, 61921006, 62006112), NSF of Jiangsu Province (BK20200313).
	We are thankful for the generous support by Ohio Supercomputer Center and AWS Cloud Credits for Research.}

{\small
\bibliographystyle{ieee_fullname}
\bibliography{main}
}

%\newpage
\clearpage
\appendix
\begin{strip}
\centering
\textbf{\Large Supplementary Material:\\ [0.2em] Procrustean Training for Imbalanced Deep Learning}
\vskip 10pt
\author{Han-Jia Ye\qquad De-Chuan Zhan\\
\small State Key Laboratory for Novel Software Technology, Nanjing University, China\\
{\tt\small \{yehj, zhandc\}@lamda.nju.edu.cn}\\
Wei-Lun Chao\\
\small The Ohio State University, USA\\
{\tt\small chao.209@osu.edu}
}
\vspace{10pt}
\end{strip}

%\appendix
%\section*{Supplementary Material}
%!TEX root=main.tex
We provide details omitted in the main paper. 
\begin{itemize}
    \item \autoref{sup-sec: statistics}: details of the statistics in Figure~1 and \autoref{fig:deviation} (cf. \autoref{s_intro} and \autoref{ss_abs} of the main paper).
    \item \autoref{sup-sec: analysis}: additional analysis of \MFW (cf. \autoref{ss_MFW_explain} of the main paper).
    \item \autoref{sup-sec: setup}: additional details of the experimental setups (cf. \autoref{sec:setup} and \autoref{ssec:mfw_imp} of the main paper).
    \item \autoref{sup-sec: results}: additional experimental results and analysis (cf. \autoref{sec:result} and \autoref{ss_abs} of the main paper).
\end{itemize}

%%%%%%%%%%%%%%%
\section{Details of the statistics of \MFW}
\label{sup-sec: statistics}

\subsection{Classification ratio} We showed in \autoref{fig:main-erm_stat} (b) of the main paper the \emph{classification ratio} per class, which is the ratio of ``the number of \emph{training} data that are classified into a class'' (\eg, class-1 or class-10) to ``the number of \emph{training} data which truly belong to that class.'' For instance, if class-10 has 50 training instances but there are 100 training instances (from all the classes) classified as class-10, then the ratio is $100/50=200~(\%)$.
Concretely, we apply the learned models (at different epochs) to classify all training data, and compute the ratio of class $c$ as follows 
\begin{equation}
\frac{\sum_{n=1}^N \textbf{1}\left[c=\argmax_{c'}\vw_{c'}^\top f_{\vtheta} (\vx_n)\right]}{N_c}.
\end{equation}
$\textbf{1}[x]$ returns 1 if $x$ is true and 0 otherwise.
This ratio offers a measure of the training progress of each class. 
If more training data are classified into a particular class than the data that class truly has, the ratio would be high, indicating that the class has been fitted more than other classes. 

\begin{figure}[t!]
	\centering
	%\minipage{0.48\linewidth}
	\centering
	\mbox{Classification ratio of ERM}
	\includegraphics[width=1.\linewidth]{figures/figure1_ERMb}\vskip 15pt
	\mbox{Classification ratio of \MFW}
	\includegraphics[width=1.\linewidth]{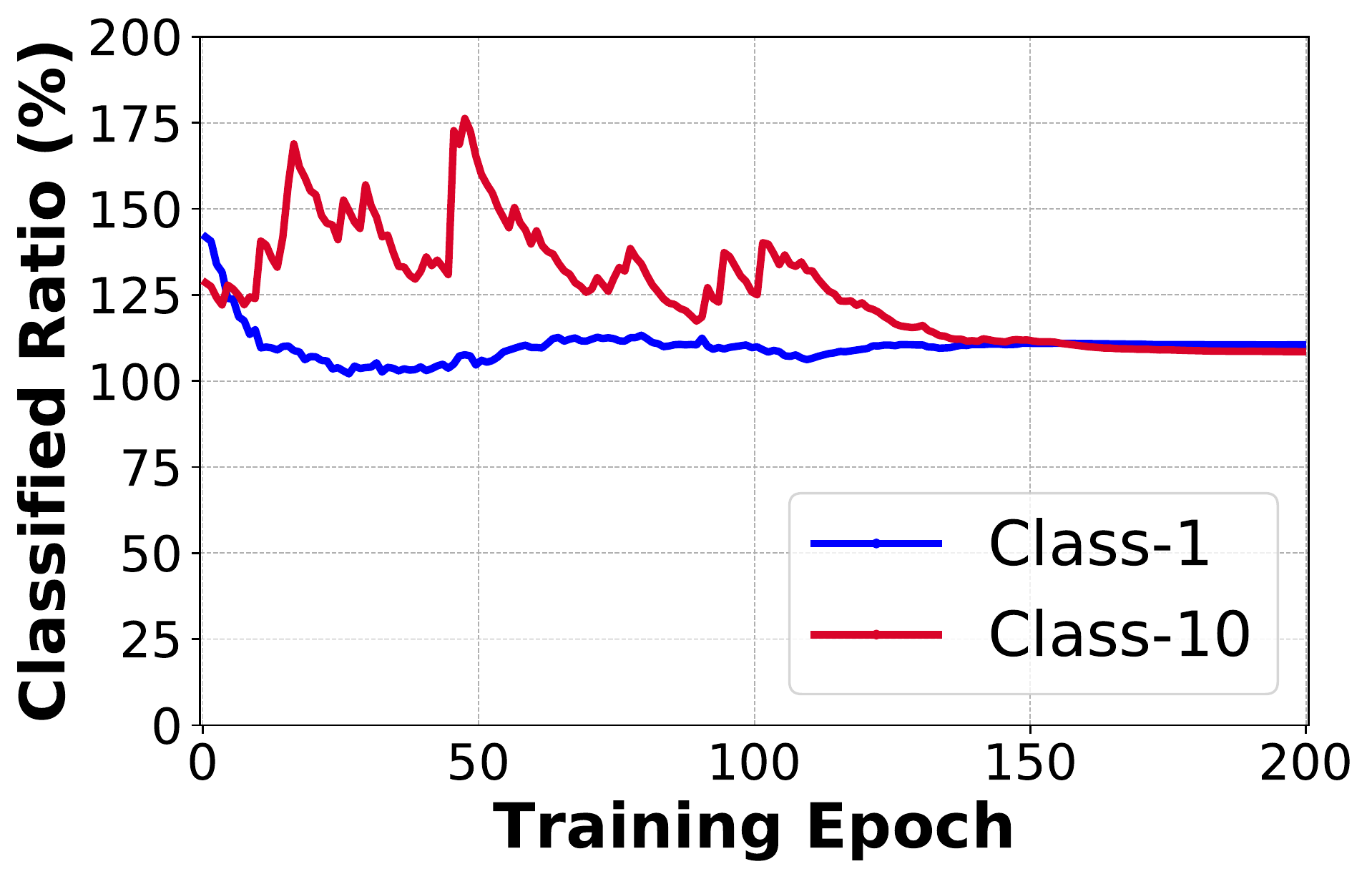}
	\caption{\textbf{The training progress of a neural network on class-imbalanced data.} We trained a ConvNet classifier using ResNet-32~\cite{he2016deep} on a long-tailed CIFAR-10 data set~\cite{krizhevsky2009learning}, following~\cite{cui2019class}. We only showed two classes for clarity. Class $c = 1$ and $c = 10$ have $5000$ and $50$ training instances, respectively. 
	The classification ratio, \ie, the numbers of training data that are classified into a class divided by the number of training data that class truly has, is plotted along the epochs for both ERM (top) and \MFW (bottom).}
	\label{fig:supp-erm_stat}
\end{figure}

In~\autoref{fig:supp-erm_stat}, we compare the classification ratio of a minor and a major class using either a conventionally-trained neural network (using ERM; top) or the one trained with our \MFW (bottom). No matter which method is used, the two classes
attain nearly $100\%$ ratio at the end of the training, which matches the fact that a neural network could ultimately fit the training data perfectly~\cite{zhang2016understanding}. 

However, if we compare the ratios along the training process, we see that using ERM, the major class has a much higher ratio (over $100\%$) in the beginning while the minor class has a much smaller ratio (around $0\%$ in the beginning). This means that most of the training data are classified into major classes at the early epochs. In contrast, with \MFW, the ratios become much balanced across classes, indicating a more balanced training progress.

\subsection{Feature deviation} 
\label{sup-subsec: statistics_FD}
We showed the formula of computing the feature deviation per class between its training and test data in \autoref{ss_abs} of the main paper.
We followed \cite{Ye2020CDT} to perform multiple rounds of sub-sampling to compute the training mean. This is to reduce the influence of the estimation variance resulting from class sizes --- even for the same major class, computing its mean using fewer data has a larger variance and therefore a larger deviation. To remove such a factor and to more faithfully reflect feature deviation across classes, we sub-sample the same number of training data per class to compute its training mean.
Similar to \cite{Ye2020CDT}, we observed a trend of increasing deviation from major to minor classes (in \autoref{fig:deviation} of the main paper), yet \MFW leads to a much smaller deviation than ERM.

%%%%%%%%%%%%%%%
\section{Analysis of \MFW}
\label{sup-sec: analysis}
\subsection{$h_{\vtheta}$ is linear}
\underline{Following \autoref{ss_MFW_explain} of the main paper}, if $h_{\vtheta}$ is linear, we have $f_{\vtheta} = \mV^\top g_{\vtheta}$. When \MFW is not performed (\ie, $\lambda_1=\lambda_2=0$), we have,
\begin{align}
\nabla_{g_{\vtheta}(\vx_1)} \ell = (\sigma(\vw^\top f_{\vtheta}(\vx_1))-y_1)\times\mV\vw, \nonumber\\
\nabla_{g_{\vtheta}(\vx_2)} \ell = (\sigma(\vw^\top f_{\vtheta}(\vx_2))-y_2)\times\mV\vw. \label{supp_e_noMFW}
\end{align}
When \MFW is performed and we apply a weight function that gives class $c=1$ a weight $0.5$ and class $c=0$ a weight $0$ (so major classes have larger weights), we have $\lambda_1\in[0, 0.5]$ while $\lambda_2=0$.
This leads to
\begin{align}
\nabla_{\tilde{\vz}_1} \ell = (\sigma(\vw^\top\mV^\top \tilde{\vz}_1)-y_1)\times\mV\vw, \nonumber\\
\nabla_{\tilde{\vz}_2} \ell = (\sigma(\vw^\top\mV^\top \tilde{\vz}_2)-y_2)\times\mV\vw, 
\end{align} 
which, by passing the gradients back to $g_{\vtheta}(\vx_1)$ and $g_{\vtheta}(\vx_2)$ (note that, we set $\lambda_2=0$ already), then gives us
\begin{align}
\nabla_{g_{\vtheta}(\vx_1)} \ell = & {\color{blue}(1-\lambda_1)} \times (\sigma(\vw^\top\mV^\top {\color{blue}\tilde{\vz}_1})-y_1)\times\mV\vw, \nonumber\\
\nabla_{g_{\vtheta}(\vx_2)} \ell = & (\sigma(\vw^\top f_{\vtheta}(\vx_2))-y_2)\times\mV\vw \nonumber \\
 & +{\color{blue}\lambda_1} \times (\sigma(\vw^\top\mV^\top {\color{blue}\tilde{\vz}_1})-y_1)\times\mV\vw. \label{supp_separate} 
\end{align} 
The second part of  $\nabla_{g_{\vtheta}(\vx_2)} \ell$ comes from $g_{\vtheta}(\vx_2)$ being used to weaken $g_{\vtheta}(\vx_1)$.
Now suppose $\vx_2$ is not classified correctly by the current model, \ie $\sigma(\vw^\top f_{\vtheta}(\vx_2))>0.5$, we have
\begin{align}
& |(\sigma(\vw^\top f_{\vtheta}(\vx_2))-y_2)| \geq \nonumber \\
& |(\sigma(\vw^\top f_{\vtheta}(\vx_2))-y_2) + {\color{blue}\lambda_1} \times (\sigma(\vw^\top\mV^\top {\color{blue}\tilde{\vz}_1})-y_1)| \geq 0, \nonumber
\end{align}
which means the norm of $\nabla_{g_{\vtheta}(\vx_2)} \ell$ will be reduced with \MFW\footnote{One can show this by plugging in $y_1 = 1$ and $y_2=0$, and consider $( \sigma(\vw^\top f_{\vtheta}(\vx_2))-0)>0.5$ and $\lambda_1 \times (\sigma(\vw^\top\mV^\top {\tilde{\vz}_1})-y_1)\in[-0.5, 0.0]$.}, compared to \autoref{supp_e_noMFW}.

\subsection{$h_{\vtheta}$ is nonlinear}
\underline{Let us define $\vz_1 = g_{\vtheta}(\vx_1)$ and $\vz_2 = g_{\vtheta}(\vx_2)$.}
When $h_{\vtheta}$ is nonlinear (\eg, by a neural network block), \autoref{supp_separate} becomes
\begin{align}
\nabla_{g_{\vtheta}(\vx_1)} \ell = & {\color{blue}(1-\lambda_1)} \times (\sigma(\vw^\top h_{\vtheta}( {\color{blue}\tilde{\vz}_1}))-y_1)\times(\nabla_{\tilde{\vz}_1}\mJ)^\top\vw, \nonumber\\
\nabla_{g_{\vtheta}(\vx_2)} \ell = & (\sigma(\vw^\top f_{\vtheta}(\vx_2))-y_2)\times(\nabla_{{\vz}_2}\mJ)^\top\vw \nonumber \\
 & + {\color{blue}\lambda_1} \times (\sigma(\vw^\top h_{\vtheta}( {\color{blue}\tilde{\vz}_1}))-y_1)\times(\nabla_{\tilde{\vz}_1}\mJ)^\top\vw, \label{supp_non_separate} 
\end{align} 
where $\nabla_{\tilde{\vz}_1}\mJ$ is the Jacobian matrix of $h_{\vtheta}(\tilde{\vz}_1)$ w.r.t. $\tilde{\vz}_1$ and $\nabla_{{\vz}_2}\mJ$ is the Jacobian matrix of $h_{\vtheta}({\vz}_2)$ w.r.t. ${\vz}_2$. Suppose that $(\nabla_{{\vz}_2}\mJ)^\top\vw$ and $(\nabla_{\tilde{\vz}_1}\mJ)^\top\vw$ are pointing to similar directions (\eg. with a high cosine similarity), then the conclusion above on gradient reduction still holds.

\subsection{Linear decision boundary $\vw$}
Still following \autoref{ss_MFW_explain} of the main paper and the notations defined above,
when there is no \MFW, the gradient to $\vw$ with the two data instances is
\begin{align}
\nabla_{\vw} \ell = & (\sigma(\vw^\top f_{\vtheta}(\vx_1))-y_1)\times f_{\vtheta}(\vx_1) \nonumber\\
 & + (\sigma(\vw^\top f_{\vtheta}(\vx_2))-y_2)\times f_{\vtheta}(\vx_2).
\end{align}

When \MFW is performed (with the same setting as above: $\lambda_1\in[0, 0.5]$ and $\lambda_2=0$), the gradient to $\vw$ becomes
\begin{align}
\nabla_{\vw} \ell = & (\sigma(\vw^\top h_{\vtheta}(\tilde{\vz}_1))-y_1)\times h_{\vtheta}(\tilde{\vz}_1) \nonumber\\
& +(\sigma(\vw^\top f_{\vtheta}(\vx_2))-y_2)\times f_{\vtheta}(\vx_2),
\end{align}
in which the first term's gradient direction changes from $f_{\vtheta}(\vx_1)=h_{\vtheta}({\vz}_1)$ to $h_{\vtheta}(\tilde{\vz}_1)$. In the case where $h_{\vtheta}$ is an identity function, this means that 
\begin{align}
\nabla_{\vw} \ell = & (\sigma(\vw^\top \tilde{\vz}_1)-y_1)\times \left((1-\lambda)g_{\vtheta}(\vx_1) + \lambda g_{\vtheta}(\vx_2)\right) \nonumber\\
& +(\sigma(\vw^\top g_{\vtheta}(\vx_2))-y_2)\times g_{\vtheta}(\vx_2).
\end{align}
In other words, \MFW also weakens the tendency of the linear classifier $\vw$ to fit the major class data\footnote{The first term now moves $\vw$ toward $(1-\lambda)g_{\vtheta}(\vx_1) + \lambda g_{\vtheta}(\vx_2)$, not $g_{\vtheta}(\vx_1)$.}. This is another reason why the overall training progress with \MFW can be more balanced across classes.

%%%%%%%%%%%%%%%
\section{Experimental Setups}
\label{sup-sec: setup}
\subsection{Datasets}
To study the imbalanced classification problems on balanced datasets (\eg, CIFAR-10~\cite{krizhevsky2009learning}, CIFAR-100~\cite{krizhevsky2009learning}, Tiny-ImageNet~\cite{le2015tiny}), we follow \cite{cao2019learning,cui2019class} to create imbalanced versions by reducing the number of training instances, such that the numbers of instances per class follow a certain distribution.
Specifically, the \emph{long-tailed} imbalance follows an exponential distribution. 
We control the degree of dataset imbalance by the imbalance ratio $\rho=\frac{N_\text{max}}{N_\text{min}}$, where $N_\text{max}$ (resp. $N_\text{min}$) is the number of training instances of the largest major (resp. smallest minor) class.

Tiered-ImageNet~\cite{Ren2018Meta} is a subset of ImageNet~\cite{deng2009imagenet} widely used in few-shot learning \cite{wang2019simpleshot}. The images are $84\times 84$. There are three splits with disjoint classes in {Tiered-ImageNet}: $351$ classes
for many-shot model training, $97$ classes for few-shot model validation, and $160$ classes for few-shot model testing.
We use {Tiered-ImageNet} to synthesize a large-scale step imbalanced dataset. 
We treat the $351$ many-shot classes as the major classes, each with around $1,000$ training images. We treat classes in the other two splits as minor classes: we randomly select $5$ training instances per class. All the classes have $3$ validation instances and $50$ test instances. We tune hyper-parameters on the $351 + 97$ classes using their validation instances: the $351$ classes are many-shot (\ie, major) and the $97$ are few-shot classes (\ie, minor). We then re-train the model with selected hyper-parameters and test it on the $351 + 160$ classes using their test instances: the $351$ classes are many-shot (\ie, major) and the $160$ are few-shot classes (\ie, minor).

In this supplementary material, we further experiment on ImageNet-LT~\cite{liu2019large}, which is also a subset of ImageNet~\cite{deng2009imagenet}. ImageNet-LT is a \emph{long-tailed} imbalance dataset with ratio $\rho=256$. It contains 1,000 classes with $115.8$K training images in total, 20 images per class for validation, and 50 images per class in the test set.

Except for the Tiered-ImageNet dataset described above, we follow \cite{cui2019class,cao2019learning,Zhou2020BBN} to report the accuracy on the test set for CIFAR-10 / CIFAR-100 / ImageNet-LT and the accuracy on the validation set for Tiny-ImageNet and iNaturalist.

\subsection{Implementation details of \MFW}
\label{supp:ssec:heldout}
For all the experiments, we use mini-batch stochastic gradient descent (SGD) with momentum = 0.9 as the optimization solver. The softmax cross-entropy loss is used for \MFW. We apply \MFW after the embedding of the second group of convolutional layers. 

We tune the beta distribution parameter $\alpha$ based on the performance on the held-out set (except for Tiered-ImageNet, which has a separate validation set from the test set). 
Concretely, we split a held-out set from the training set, following \cite{Ye2020CDT}.  We  held  out 3  images  per class, creating a small balanced held-out set. If a class has fewer than 3 images, we ignore them in hyper-parameter tuning.

We keep the scale value in \autoref{eq:lambda_n} of the main paper as $\beta=2$ and $\beta=0.01$ for the long-tailed and step cases, respectively. Deferred Re-Weight~(DRW) \cite{cao2019learning} is also applied to \MFW near the end of the training process, to further improve the accuracy. In detail, we apply the vanilla softmax (with \MFW) at first, and after completing 80\% of the training progress, we apply a weighted version of softmax for optimization. The class-wise weights are set based on~\cite{cui2019class}.

\subsection{Training details for imbalanced CIFAR}
We use ResNet-32~\cite{he2016deep} for all the  CIFAR~\cite{krizhevsky2009learning} experiments, following~\cite{cao2019learning, cui2019class}. The batch size is 128, and the weight decay is $2 \times 10^{-4}$. The initial learning rate is linearly warmed up to $0.1$ in the first five epochs, and decays with cosine annealing. The model is trained for 300 epochs\footnote{On CIFAR, we found that for ERM and other baselines, training with 200 epochs converges. As MFW weakens major features, some cases need more epochs to converge and we train them for 300.}.
We follow \cite{cao2019learning} to do data augmentation. The $32 \times 32$ CIFAR images are padded to $40 \times 40$ and randomly flipped horizontally, and then are randomly cropped back to $32 \times 32$.  

\begin{figure*}[t!]
	\begin{minipage}{0.245\linewidth}
		\centering
		\mbox{Long-tailed (ERM)}
		\includegraphics[width=\linewidth]{figures/figure5_exp_deviation_ERM.pdf}
	\end{minipage}
	\begin{minipage}{0.245\linewidth}
		\centering
		\mbox{Long-tailed (\MFW)}
		\includegraphics[width=\linewidth]{figures/figure5_exp_deviation_Mix.pdf}
	\end{minipage}
	\begin{minipage}{0.245\linewidth}
		\centering
		\mbox{Step (ERM)}
		\includegraphics[width=\linewidth]{figures/figure5_step_deviation_ERM.pdf}
	\end{minipage}
	\begin{minipage}{0.245\linewidth}
		\centering
		\mbox{Step (\MFW)}
		\includegraphics[width=\linewidth]{figures/figure5_step_deviation_Mix.pdf}
	\end{minipage}
	
	\begin{minipage}{1.01\linewidth}
		\centering
		% \mbox{Step (\MFW)}
		\includegraphics[width=\linewidth]{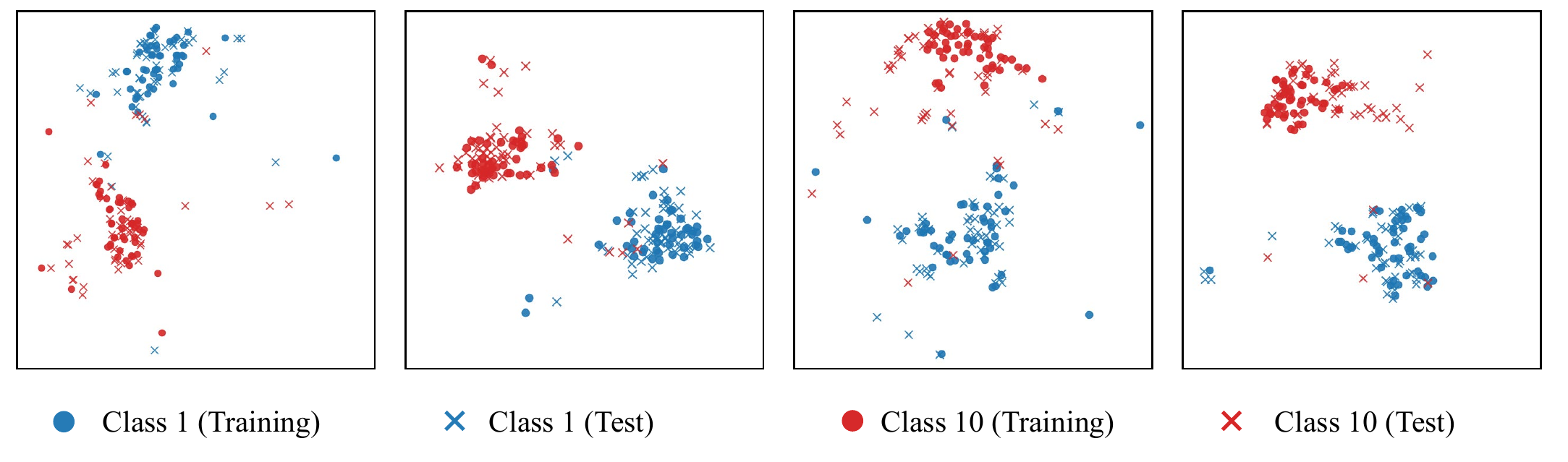}
	\end{minipage}
	\caption{\textbf{Feature deviation between the training and test data per class along the training progress.} We experiment on CIFAR-10, using both the long-tailed and the step settings ($\rho=100$).
	We only showed the most major ($c = 1$, with $5,000$ training samples) and minor classes ($c = 10$, with $50$ training samples) for clarity.
	As the number of training epochs increases, the deviation increases, while \MFW can achieve a much smaller deviation (top row). The bottom parts are the corresponding t-SNE embeddings of the training and test instances from the two classes (at the last epoch).}
	\label{supp-fig:deviation}
\end{figure*}

\subsection{Training details for Tiny-ImageNet}
We use ResNet-18~\cite{he2016deep}, following~\cite{cao2019learning}. It is trained for 200 epochs with a batch size of 128. The initial learning rate is 0.1 and decays with cosine annealing. Images are padded 8 pixels on each size and randomly flipped horizontally, and then are randomly cropped back to $64 \times 64$.

\subsection{Training details for Tiered-ImageNet}

We use ResNet-12~\cite{Lee2019Meta,Ye2020Few}. It is trained for 180 epochs with a batch size of 512, following~\cite{kang2019decoupling}. The initial learning rate is 0.2 and decays with cosine annealing. Images are padded 8 pixels on each size and randomly flipped horizontally, and then are randomly cropped back to $84 \times 84$.

\subsection{Training details for ImageNet-LT}

We use ResNet-10~\cite{he2016deep} and ResNext-50~\cite{Xie2017Aggregated}. They are trained for 200 epochs with a batch size of 256. The initial learning rate is 0.1 and decays with cosine annealing. In training, the images are resized to $256 \times 256$ and flipped horizontally, and are randomly cropped back to $224 \times 224$.

\subsection{Training details for iNaturalist} 
We follow~\cite{cui2019class,Zhou2020BBN} to use ResNet-50~\cite{he2016deep} on iNaturalist. We train the model for 90 and 180 epochs with a batch size of 128. 
The learning rate is 0.05 at first. It decays at the 60th (resp. 120th) and the 80th (resp. 160th) by 0.1 when training for 90 epochs (resp. 180 epochs).
We follow \cite{Zhou2020BBN} to apply the standard pre-processing and data augmentation used for ImageNet \cite{he2016deep}. We normalize the images by subtracting the RGB means computed on the training set. In training, the images are resized to $256 \times 256$ and flipped horizontally, and are randomly cropped back to $224 \times 224$.

%%%%%%%%%%%%%%%%%%%%
\section{Experimental Results and Analysis}
\label{sup-sec: results}
We provide additional experimental results and analysis in this section.

\subsection{The influence of the coefficient $\alpha$}
According to Algorithm \ref{a_MFW} of the main paper, we sample a value $\lambda_n$ from a beta distribution with a coefficient $\alpha$ to mix two features in \MFW. With different $\alpha$ values, the beta distribution behaves differently. For example, the beta distribution with small $\alpha$ values (\ie, $\alpha<1$) favors sampling extreme $\lambda_n$ values close to 0 or 1; large $\alpha$ increases the probability of sampling values near 0.5. We show the results when using different $\alpha$ in~\autoref{tab:mix_alpha}. We find that for the step case, a larger $\alpha$ is preferred; for the long-tailed case, a smaller $\alpha$ is preferred.
We select $\alpha$ using the held-out set (or the validation set for Tiered-ImageNet). See \autoref{supp:ssec:heldout} for details.

\begin{table}[t]
	\centering
	\caption{Test accuracy (\%) on CIFAR-10/-100 with different values of $\alpha$ (for the beta distribution) in \MFW.}
	\begin{tabular}{c|cc||cc}
		\addlinespace
		\toprule
		& \multicolumn{ 2}{c||}{CIFAR-10}                 & \multicolumn{ 2}{c}{CIFAR-100}                \\
		\midrule
		$\alpha$ & Long-tailed & Step & Long-tailed & Step \\
		\midrule
		0.1  & 75.9 & 73.3 & 43.1 & 40.8 \\
		0.2  & 76.8 & 74.3 & 44.7 & 42.6 \\
		0.5  & 77.1 & 75.2 & 43.6 & 44.0 \\
		1  & 78.5 & 77.6 & 42.1 & 46.5 \\
		5  & 77.4 & 80.1 & 41.2 & 46.9 \\
		\bottomrule
	\end{tabular}
	\label{tab:mix_alpha}
\end{table}

\begin{figure}
\centering
\includegraphics[width=0.6\linewidth]{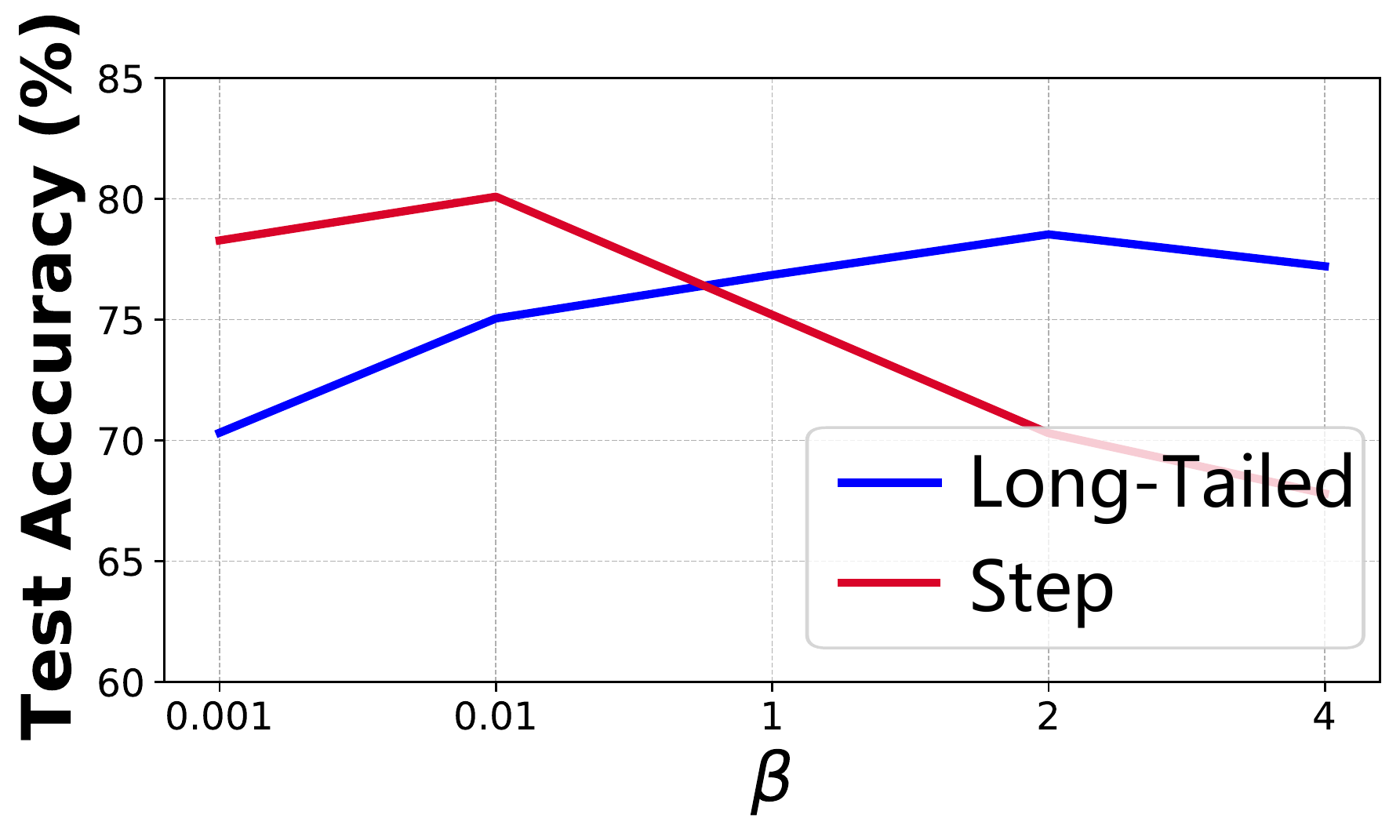}
\caption{\small Ablation on the weight function.}
\label{fig:beta}
\end{figure}

\subsection{Ablation on the weight function.} We study different $\beta$ in \autoref{eq:lambda_n} of the main paper. As shown in \autoref{fig:beta}, larger/smaller $\beta$ (smoother/sharper weight changes) is preferred for long-tailed/step cases. We further apply \autoref{eq:lambda_n} to MixUp~\cite{zhang2018mixup} and ReMix~\cite{chou2020remix} but do not see improvements. We attribute this to the fundamental difference between MFW and them (cf. \autoref{ssec:cmp_mixup} in the main paper): MFW does not change the labels, while MixUp does and ReMix mixes labels to favor minor classes.

\subsection{Whether \MFW can reduce the feature deviation?} Following \autoref{sec:result} of the main paper, we further show the t-SNE~\cite{maaten2009visualizing} plots of the features to illustrate the feature deviation in \autoref{supp-fig:deviation}. We apply ERM and \MFW on different imbalance configurations, including both long-tailed and step cases. The largest major class 1 and the smallest class 10 are selected for illustration. 

\begin{figure}[t!]
	\centering
	%\minipage{0.48\linewidth}
	\centering
	\mbox{Classification Accuracy of ERM}
	\includegraphics[width=1.\linewidth]{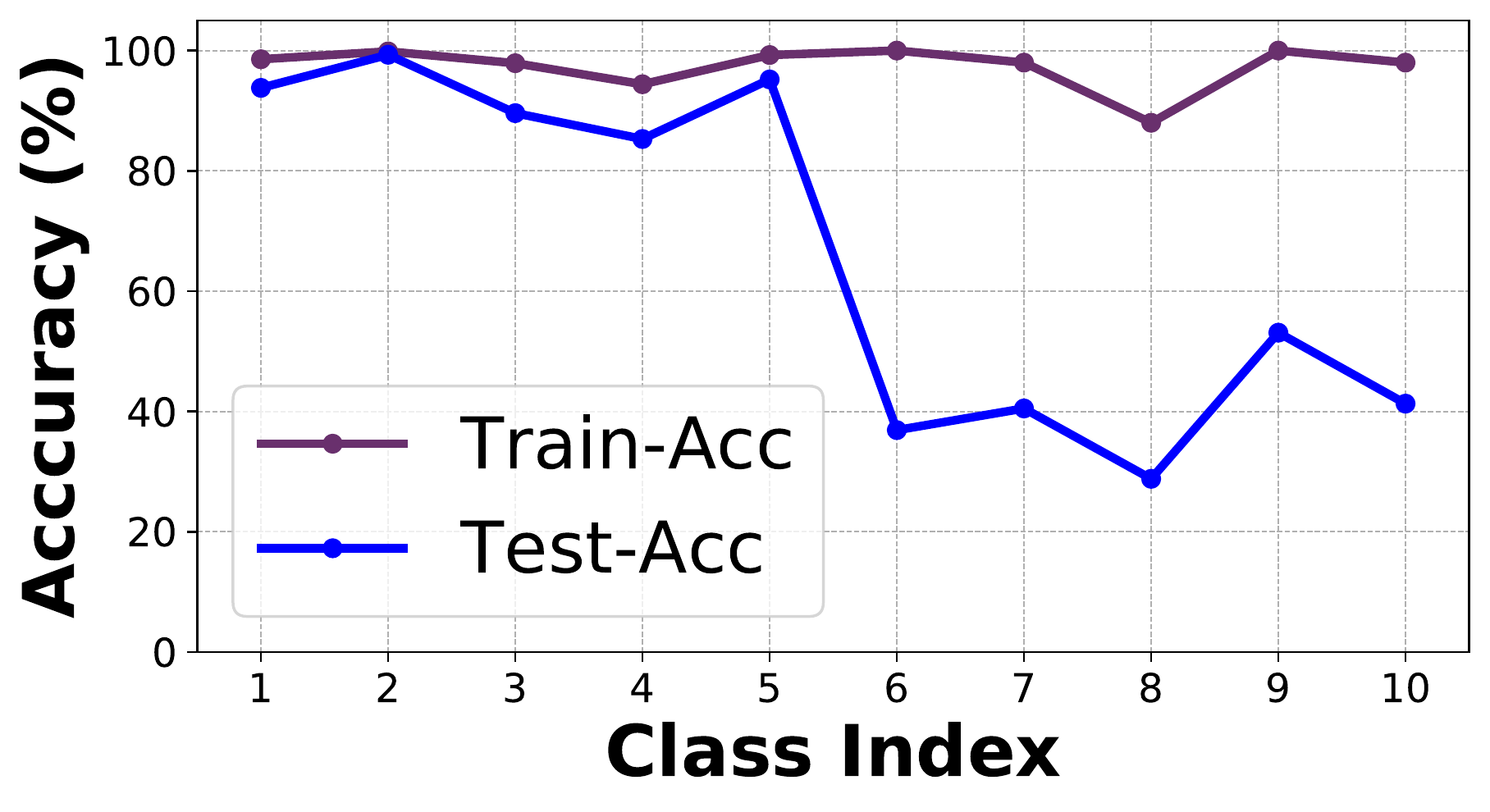}
	\mbox{Classification Accuracy of \MFW}
	\includegraphics[width=1.\linewidth]{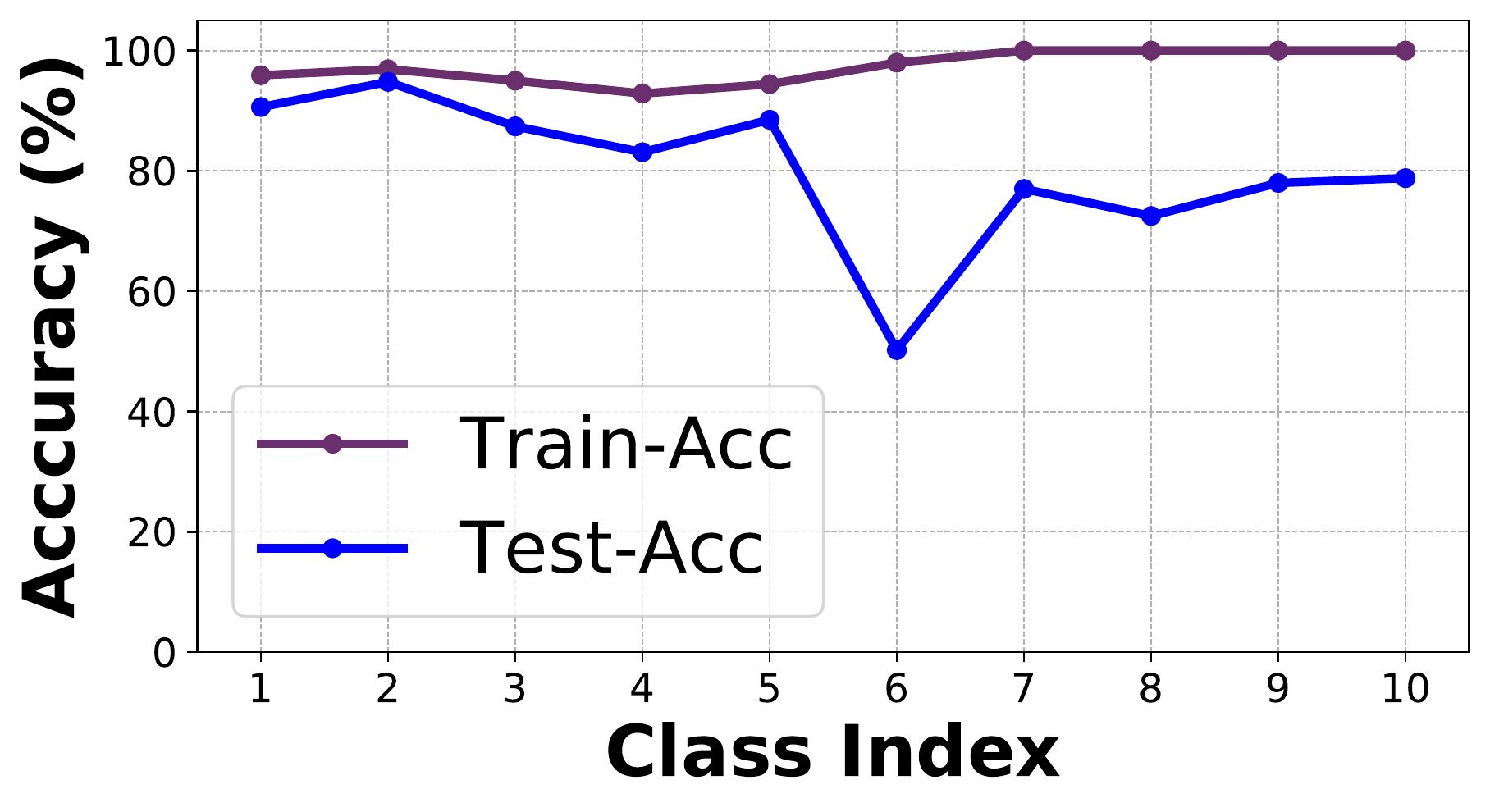}
	\caption{\textbf{The training and test accuracy per class of a neural network trained on class-imbalanced data.} We trained a ConvNet classifier using ResNet-32~\cite{he2016deep} on a step imbalanced CIFAR-10 data set ($\rho=100$)~\cite{krizhevsky2009learning}, following~\cite{cui2019class}. The first five and last five Classes have $5,000$ and $50$ training instances, respectively. 
	We compare training with ERM (top) and \MFW (bottom). \MFW leads to much higher test accuracy for the minor classes, with just a slight decrease of that for the major classes.}
	\label{fig:supp-per-class-acc}
\end{figure}

In \autoref{supp-fig:deviation}, the top row shows the change of feature deviation values\footnote{Please see \autoref{ss_abs} of the main paper and \autoref{sup-subsec: statistics_FD} for the definition.} along the training process, where \MFW significantly reduces the feature deviation, justifying our claims that feature deviation is likely caused by the exaggerated gradients between major and minor classes. The bottom row shows the t-SNE visualization of $f_{\vtheta}(\vx)$ using the corresponding model (at the last epoch): circles indicate the training data and crosses indicate the test data. It can be seen that there exists feature deviation for class 10 (red) when applying ERM, which can be reduced using \MFW. With \MFW, the training and test features are clustered.

\subsection{Training and test accuracy across classes} We plot the training and test accuracy on the step imbalanced CIFAR-10 ($\rho=100$) in \autoref{fig:supp-per-class-acc}. We note that, in evaluating the training accuracy, we do not apply \MFW --- \MFW is only applied in training the neural networks. We mention this in \autoref{ss_MFW} and \autoref{ss_balanced_progress} in the main paper.

As shown in \autoref{fig:supp-per-class-acc}, \MFW leads to a much higher test accuracy for the minor classes (the average is over $60\%$ in comparison to $40\%$ by ERM), with just a slight decrease of that for the major classes. In summary, \MFW can effectively facilitate class-imbalanced learning, especially for the challenging step imbalance cases.

\begin{table}[t]
	\centering
	\small
	\caption{Test set accuracy (\%) on imbalanced CIFAR-10 with larger imbalance ratio $\rho$.}
	\begin{tabular}{c|cc}
		\addlinespace
		\toprule
		Imbalance ratio $\rho$ & 500 & 1000 \\
		\midrule
		ERM   & 64.5  & 56.5 \\
		CDT~\cite{Ye2020CDT}   & 66.1  & 59.0 \\
		\midrule
		MFW   & 65.6  & 62.0 \\
		MFW w/ DRW & 67.3  & 63.0 \\
		\bottomrule
	\end{tabular}
	\label{tab:large_rho}
\end{table}

\subsection{Label mixing} We compare \MFW to mixup \cite{zhang2018mixup} and manifold mixup \cite{verma2019manifold}: they mix both the inputs (or features) and the labels.  As shown in \autoref{tab:cifar-abltaion-suppl}, \MFW outperforms both of them, justifying that \MFW is not regularizing the model predictions.

\begin{table}[t]
	\small
	\centering
	\tabcolsep 3pt
	\caption{\small Test accuracy (\%) on CIFAR-10/-100. We compare \MFW to \cite{zhang2018mixup,verma2019manifold}.}
	\vskip-2pt
	\begin{tabular}{l|cc||cc}
		\addlinespace
		\toprule
		& \multicolumn{ 2}{c||}{CIFAR-10}                 & \multicolumn{ 2}{c}{CIFAR-100}                \\
		\midrule
		Imbalance ratio $\rho=100$ & Long-tailed & Step & Long-tailed & Step \\
		\midrule
		ERM & 71.1 & 65.8 & 40.1 & 39.9 \\ %\hline
		Mix-Up~\cite{zhang2018mixup} & 73.1 & 66.2 & 39.5 & 39.8 \\
		Manifold Mix-up~\cite{verma2019manifold} & 73.0 & 65.4 & 38.3 & 39.4\\
		%\midrule
		{\MFW} & \textbf{78.5} & \textbf{80.1}  & \textbf{44.7} & \textbf{46.9} \\
		\bottomrule
	\end{tabular}
	\label{tab:cifar-abltaion-suppl}
\end{table}

\subsection{Performance on larger imbalance factors}
We study the performance of \MFW on a long-tailed scenario with larger imbalance ratio $\rho$ on CIFAR-10 in \autoref{tab:large_rho}.
In detail, we construct the dataset in the same manner as~\cite{cui2019class,cao2019learning} and set $\rho=500$ and $\rho=1000$.
Since the dataset is more imbalanced, it becomes more difficult. 
Compared to a strong baseline CDT~\cite{Ye2020CDT}, the gain by MFW increases as $\rho$ increases. Overall, we find that \MFW has advantages in tackling extreme imbalance, \eg, step settings (all minor classes have $\frac{1}{\rho}$ training data than the major classes) or larger $\rho$ in long-tailed settings.

\begin{table}[t]
	\small
	\centering
	\caption{Top-1 test set accuracy (\%) on ImageNet-LT.}
	\begin{tabular}{c|c||c|c}
		\addlinespace
		\toprule
		\multicolumn{ 2}{c||}{ResNet-10} & \multicolumn{ 2}{c}{ResNeXt-50} \\
		\midrule
		OLTR~\cite{liu2019large} & 41.9  & OLTR~\cite{liu2019large}  & 48.7 \\
		cRT~\cite{kang2019decoupling}   & 41.8  & cRT~\cite{kang2019decoupling}   & 49.6 \\
		$\tau$-Norm~\cite{kang2019decoupling} & 40.6  & $\tau$-Norm~\cite{kang2019decoupling} & 49.4 \\
		LWS~\cite{kang2019decoupling}   & 41.4  & LWS~\cite{kang2019decoupling}   & 49.9 \\
		CDT~\cite{Ye2020CDT}   & 41.4  & De-confound~\cite{tang2020long} & 48.6 \\
		SEQL~\cite{tan2020equalization}  & 36.4  & De-confound-TDE~\cite{tang2020long} & 51.8 \\
		\midrule
		MFW   & 40.1  & MFW   & 50.5 \\
		MFW w/ DRW & 42.0    & MFW w/ DRW & 51.9 \\
		\bottomrule
	\end{tabular}
	\label{tab:imagenet}
\end{table}

\subsection{Results on ImageNet-LT}
We show the top-1 accuracy on ImageNet-LT in \autoref{tab:imagenet} with ResNet-10 and ResNext-50. Our \MFW achieves promising results when compared with others.

\subsection{Further comparisons on CIFAR}
\cite{tan2020equalization} mentioned that they used stronger data augmentations on CIFAR. Using the same augmentations, we have $47.8$ on CIFAR-100 ($\rho$=200) vs. theirs at $43.4$. 

\subsection{Deferred Re-Weight~(DRW)} For step imbalance, MFW w/o DRW outperforms baselines in nearly all cases. (See tables in the main paper.) We include DRW mainly for a fair comparison to~\cite{chou2020remix}. We note that, deferred scheduling (\eg, DRW) is also included in the implementation of other methods, \eg, \cite{cao2019learning,Jamal2020Rethinking,kang2019decoupling,Kim2020M2M,Zhou2020BBN}.

\subsection{\MFW on other network architectures} We further study the generalizability of \MFW to other network architectures. Using DenseNet-121~\cite{huang2017densely}, \MFW achieves $76.2/81.2/92.1$ for long-tailed CIFAR-10 at $\rho=200/100/10$, higher than ERM which has $70.2/77.6/91.3$ (cf. \autoref{tab:cifar-table} in the main paper).

\end{document}